\pdfoutput=1

\documentclass[11pt]{article}
\usepackage{acl}

\usepackage{times}
\usepackage{latexsym}
\usepackage[T1]{fontenc}
\usepackage[utf8]{inputenc}
\usepackage{microtype}
\usepackage{inconsolata}

\usepackage{chngcntr}  
\usepackage{amsmath}
\usepackage{amssymb}
\usepackage{graphicx}
\usepackage{booktabs}
\usepackage{tabularx}
\usepackage{multirow}
\usepackage{siunitx}  
\usepackage{etoolbox}
\newrobustcmd\B{\DeclareFontSeriesDefault[rm]{bf}{b}\bfseries}

\usepackage[capitalize,nameinlink,noabbrev]{cleveref}
\creflabelformat{equation}{#2\textup{#1}#3}
\crefname{section}{\S}{\S\S}  
\crefname{subsection}{\S}{\S\S}  
\crefname{subsubsection}{\S}{\S\S} 

\crefname{codeboxinput}{Listing}{Listings}
\usepackage[shortcuts]{extdash}  
\usepackage{subcaption}
\usepackage{fontawesome5}
\usepackage{algorithm}
\usepackage{algpseudocodex}

\usepackage{enumitem}
\newlist{todolist}{itemize}{2}
\setlist[todolist]{label=$\square$}

\usepackage{soul}
\definecolor{pastelblue}{HTML}{A1C9F4}
\definecolor{pastelorange}{HTML}{FFB482}
\definecolor{pastelgreen}{HTML}{8DE5A1}
\definecolor{pastelred}{HTML}{FF9F9B}
\definecolor{pastelpurple}{HTML}{D0BBFF}

\newcommand{\hlorange}[1]{\sethlcolor{pastelorange}\hl{#1}}

\newcommand{\hlgreen}[1]{\sethlcolor{pastelgreen}\hl{#1}}

\usepackage[most]{tcolorbox}
\tcbset {
  base/.style={
    arc=0mm,
    bottomtitle=0.5mm,
    boxrule=0mm,
    colbacktitle=black!10!white,
    colframe=black!30!white,
    coltitle=black,
    fonttitle=\bfseries\fontsize{10}{12}\selectfont,
    left=2.5mm,
    leftrule=1mm,
    right=3.5mm,
    title={#1},
    toptitle=0.75mm,
    breakable,
    listing options={style=tcblatex, breakindent=0pt, breaklines=true, basicstyle=\ttfamily\fontsize{8}{9}\selectfont}
  }
}

\newtcblisting[auto counter]{codebox}[2][]{
  base={Listing \thetcbcounter: #2},
  listing only,
  #1
}

\newtcbinputlisting[use counter=codeboxinput]{\codeboxinput}[3][]{%
  listing file={#3},
  base={Listing \thetcbcounter: #2},
  listing only,
  #1
}

\newtcolorbox{subbox}[2][]{
  colframe=black!30!white,
  base={#2},
  #1
}

\newcommand\rurl[1]{%
  \href{https://#1}{\nolinkurl{#1}}%
}

\title{Behavioral Analysis of Information Salience in Large Language Models}

\author{
Jan Trienes$^{1}$\quad
Jörg Schlötterer$^{1,2}$\quad
{\bf Junyi Jessy Li}$^3$\quad
{\bf Christin Seifert}$^1$\\
$^1$Marburg University\quad
$^2$University of Mannheim\\
$^3$The University of Texas at Austin\\
\texttt{\normalsize \{jan.trienes,joerg.schloetterer,christin.seifert\}@uni-marburg.de}\\
\texttt{\normalsize jessy@utexas.edu}}

\def\gpt{GPT\=/4o}
\def\gptmini{GPT\=/4o\=/mini}

\newcommand\unfootnote[1]{%
  \begingroup
  \renewcommand\thefootnote{}\footnote{#1}%
  \addtocounter{footnote}{-1}%
  \endgroup
}

\begin{document}
\maketitle
\begin{abstract}
Large Language Models (LLMs) excel at text summarization, a task that requires models to select content based on its importance.
However, the exact notion of salience that LLMs have internalized remains unclear.
To bridge this gap, we introduce an explainable framework to systematically derive and investigate information salience in LLMs through their summarization behavior.
Using length-controlled summarization as a behavioral probe into the content selection process, and tracing the answerability of Questions Under Discussion throughout, we derive a proxy for how models prioritize information.
Our experiments on 13 models across four datasets reveal that LLMs have a nuanced, hierarchical notion of salience, generally consistent across model families and sizes.
While models show highly consistent behavior and hence salience patterns, this notion of salience cannot be accessed through introspection, and only weakly correlates with human perceptions of information salience.\footnote{We release code, model outputs and human annotations at \url{https://github.com/jantrienes/llm-salience}.}
\end{abstract}

\section{Introduction}
Large Language Models (LLMs) significantly advanced text synthesis tasks, including text summarization, which they perform well even under zero-shot conditions~\cite{Goyal:2023:arXiv,Zhang:2024:TACL}.
The nature of the summarization task requires models to do content selection: picking the most salient pieces of information for inclusion in a summary \cite{Mani:1999:advances}.
However, it remains unclear what underlying notion of salience the models have internalized.

Prior work investigated information salience from several angles.
Theories of discourse structure have been used to induce content salience \cite{Marcu:1999:advances,Louis:2010:SIGDIAL}, and a large body of summarization research uses word distribution or centrality as the main signal for content selection \cite{Nenkova:2012:survey,Nazari:2019:survey}. \citet{Peyrard:2019:ACL} laid out a theoretical perspective for content salience in summarization, though the exact notion remains largely latent and aloof; rather, pre-LLM summarization work uses human summaries as supervision signals to learn what to include \cite[][\emph{inter alia}]{Gehrmann:2018:EMNLP,Chen:2018:ACL,Liu:2019:EMNLP}.
Yet, none of these accounts explains why LLM zero-shot summarization works so well on the one hand, while missing key elements on the other~\cite{Kim:2024:COLM,Trienes:2024:ACL,Huang:2024:NAACL}.

To begin to make sense of this behavior, we need to understand how models internalize salience: whether it is a \emph{consistent} notion within and across models, \emph{how} they prioritize information, and whether LLMs' notion of salience \emph{aligns} with prior theories or human intuitions.

In this paper, we present a novel explainable framework
to systematically derive and investigate LLMs' grasp of information salience through their summarization behavior.
Our method combines \textbf{two key ideas}. First, we can use length-constrained summarization~\cite{Fan:2018:NGT,He:2022:EMNLP} as a behavioral probe into the content selection process of LLMs.
Intuitively, when there is a limited length budget for a summary, we posit that the least important information is dropped first.

\begin{figure*}[t]
\includegraphics[width=\textwidth]{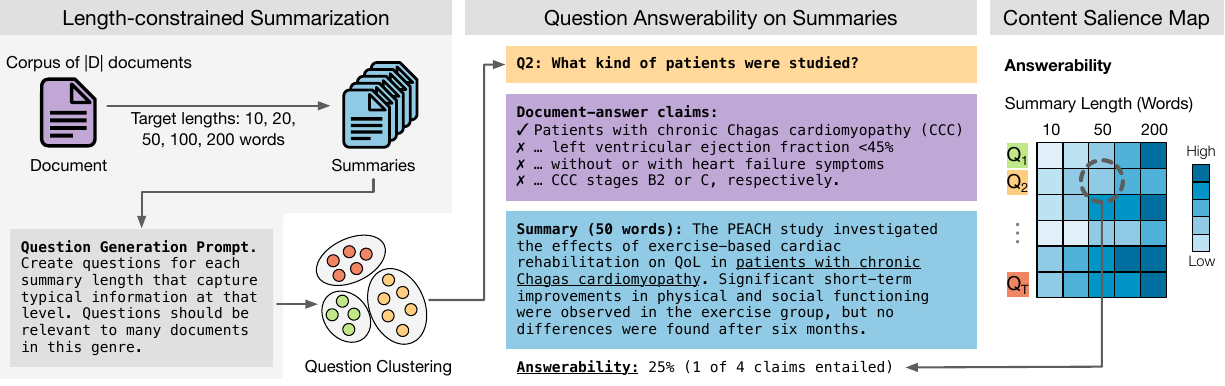}
\caption{
    Framework overview, conceptualizing content salience as question answerability.
    \textbf{Left:} Given a corpus, we derive questions that are typically answered in summaries. Length-controlled summarization acts as a probe into the content-selection process of LLMs.
    Question paraphrases are clustered by semantic intent.
    \textbf{Middle:} Answerability is calculated as the fraction of document-answer claims entailed by the summary.
    \textbf{Right:} The content salience map tracks answerability at each summary length. More salient questions remain answerable even in shorter summaries.
}
\label{fig:overview}
\end{figure*}

Second, we can describe what is salient as the answerability of domain-relevant Questions Under Discussion (QUDs;~\citealp{Van:1995:discourse,Benz:2017:questions,Wu:2023:EMNLP}). QUDs can be thought of as representations of a coherent unit of information in the form of information-seeking questions, e.g., \emph{Who are the participants of this study?}
We use such questions --- and hence their answers extracted from documents, according to alternative semantics \cite{Hamblin:1973:Questions,Karttunen:1977:syntax,Groenendijk:1984:studies} --- as the primary unit of analysis, making our framework interpretable and customizable.

Taken together, by gradually decreasing the length budget available for a summary and by systematically tracing question answerability throughout, we can derive a \emph{proxy} for how models prioritize information.
See~\cref{fig:overview} for an overview.

Using this framework (\cref{sec:method}), we empirically study LLMs' content selection \emph{behavior}, and its alignment with \emph{perceived} notions of salience. Through experiments on 13 models and four datasets (\cref{sec:experimental-settings}), we aim to answer the following research questions:
\begin{itemize}[noitemsep,leftmargin=27pt]
    \item[\textbf{RQ1}] What notion of salience have LLMs learned in different domains?
    \item[\textbf{RQ2}] Do LLMs of different families/sizes have a similar notion of salience?
    \item[\textbf{RQ3}] When models introspect, does their perceived notion of salience align with their summarization behavior?
    \item[\textbf{RQ4}] To what extent does model salience align with human perceived salience?
\end{itemize}
We find that LLMs have a nuanced notion of salience prioritizing information hierarchically across summary lengths (\cref{sec:results-salience}).
Also the notion of salience is generally compatible between models even of different families and sizes, though more recent/bigger LLMs correlate more strongly with GPT-4o (\cref{sec:results-model-model-similarity}).
Furthermore, models show highly consistent behavior and hence notions of salience, but it cannot be elicited through introspection (i.e., directly prompting for salience of topics; \cref{sec:results-introspection}). Lastly, we find that model behavior only weakly aligns with human perceptions of salience (\cref{sec:results-human-alignment}).

\section{Method: Analyzing Content Salience}
\label{sec:method}
To analyze content salience we need a way to both \emph{observe} what content models consider important (\cref{sec:method-probe}), and to \emph{describe} it in an interpretable manner (\cref{sec:method-questions}).
\cref{fig:overview} illustrates the framework.

\subsection{Length-constrained Summarization as a Content Salience Probe}
\label{sec:method-probe}
To elicit content-selection decisions from models, we use length-constrained summarization as a probe. Our key intuition is that, under a limited length budget, well-behaving models will drop the least important information first, while preserving the most salient content.

\paragraph{Summary Generation.} 
Given a corpus $D$, and a set of target lengths $L$ specified in words, we generate summaries $S = \{s_{d,l} \mid d \in D, l \in L \}$ for all documents and length targets.
We consider $L = \{10, 20, 50, 100, 200\}$ to capture a range of typical summary lengths.

\paragraph{Tracing Content-selection Decisions.}
To understand how summary content changes with varying length budgets, we introduce \textit{Content Salience Maps (CSMs)} as a structured representation to systematically track the inclusion and exclusion of topics.
Formally, let $T$ be a set of topics of interest, and let $f : T \times S \rightarrow [0,1]$ be a function that measures to what extent topic $t$ is present in summary $s$.
For a document $d$, $\text{CSM}(d)$ is a $|T|\times|L|$ matrix, where each entry is defined as:
\begin{equation}
    \text{CSM}(d)_{t,l} = f(t, s_{d,l}).
\end{equation}
We define the corpus-level $\text{CSM}(D)$ as the average of document-level measurements:
\vspace{-0.1em}%
\begin{equation}\label{eqn:corpus-level-csm}
    \text{CSM}(D)_{t,l} = \frac{1}{|D^t|} \sum\limits_{d \in D^t}f(t,s_{d,l}),
\end{equation}\vspace{-0.1em}
where $D^t$ is the set of documents that contain topic $t$.
We also define \emph{topic prevalence} as $|D^t|/|D|$, representing the fraction of documents in the corpus that contain topic $t$.

Below, we describe a concrete instantiation of this framework, where the set of topics $T$ is represented as QUDs, and the inclusion measure $f$ is defined as question answerability.
However, we note that the framework is highly customizable
in terms of the definitions of $T$ and $f$.

\subsection{Question-based Content Analysis}
\label{sec:method-questions}
We represent the topics $t \in T$ as Questions Under Discussion (QUD), a linguistic representation for topics in discourse \cite{Van:1995:discourse}.
In our setup, each QUD represents a possible \emph{answer space} across different documents in the same genre. This aligns with alternative semantics, where questions are viewed as the set of possible answers \cite{Hamblin:1973:Questions,Karttunen:1977:syntax,Groenendijk:1984:studies}.
In addition to the interpretability provided by natural language questions,
we can also quantify content salience through \emph{question answerability}:
questions which remain answerable even with shorter summaries are more salient than questions which can only be answered with longer summaries.
Below (also \cref{alg:csm-derivation}), we describe a four-step pipeline to implement this approach.

\setlength{\textfloatsep}{0.7em}
\algrenewcommand\algorithmicrequire{\textbf{Input:}}
\algrenewcommand\algorithmicensure{\textbf{Output:}}
\begin{algorithm}[t]
\captionsetup{font=10pt}
\footnotesize
\caption{Content Salience Map (CSM) Derivation}\label{alg:csm-derivation}
\begin{algorithmic}[1]
\Require Corpus: $D = \{d_1, d_2, ..., d_{|D|}\}$\\
         Lengths: $L = \{10, 20, 50, 100, 200\}$ (words)\\
         Models: $M_{\text{Sum}}, M_{\text{QG}}, M_{\text{Emb}}, M_{\text{QA}}, M_{\text{ClaimSplit}}, M_{\text{NLI}}$
\Ensure Corpus-level $\text{CSM}_D$ 
\For{$(d, l) \in D \times L$} \Comment{Step 0: Summarization}
    \State $S[d,l] \gets M_{\text{Sum}}(d, l)$
\EndFor
\Statex \vspace{-0.5em}
\For{$d \in D$} \Comment{Step 1: Question Generation}
    \State $Q \gets Q \cup M_{\text{QG}}(d, S[d,:])$
\EndFor
\Statex \vspace{-0.4em}
\State $T \gets \text{Cluster}(M_{\text{Emb}}(Q))$ \Comment{Step 2: Question Clustering}
\State $T \gets \text{ManualReview}(T)$
\State $T \gets \text{SelectClusterRepresentatives}(T)$\\[-5pt]
\For{$(d, t) \in D \times T$} \Comment{Step 3: QA and Claim Split}
\State $\text{ans}_{\text{ref}} \gets M_{\text{QA}}(d, t)$
\State \algorithmicif\ $\text{ans}_{\text{ref}} \neq \varnothing$ \algorithmicthen\ $A[d,t] \gets M_{\text{ClaimSplit}}(\text{ans}_{\text{ref}})$
    \State \algorithmicelse\ $A[d,t] \gets \varnothing$
\EndFor
\Statex \vspace{-0.5em}
\For{$(t, l) \in T \times L$} \Comment{Step 4: Answerability}
    \For{$d \in D$}
        \State $s, A_t \gets S[d,l], A[d,t]$ \Comment{(summary, claims)}
        \State $\text{CSM}_d[t,l] \gets \text{avg}([M_{\text{NLI}}(a,s) \mid a \in A_t ])$ \Comment{Eq.~\ref{eqn:answerability}}
    \EndFor
    \State $D_t \gets \{d \in D \land A[d,t] \neq \varnothing\}$
    \State $\text{CSM}_D[t,l] \gets \text{avg}([\text{CSM}_d[t,l] \mid d \in D_t])$ \Comment{Eq.~\ref{eqn:corpus-level-csm}}
\EndFor

\State \Return $\text{CSM}_D$
\end{algorithmic}
\end{algorithm}

\paragraph{Step 1: Question Generation.}
We
design a question-generation prompt inspired by \cite{Laban:2022:Findings}.
Given summaries of varying lengths from a random sample of documents, we prompt an
LLM to generate $n$ questions which each summary answers in a unique way.
The prompt specifies two requirements: (1) the questions should be answerable by most documents in the given genre, and (2) they should highlight meaningful differences between summaries of different lengths
(full prompt in \cref{sec:appendix-prompts}).
For example, in movie reviews, most summaries will answer questions such as \emph{``What is the main plot of the movie?''},
but naturally the answers will be different for each review.
We repeat this process for all documents and associated summaries in the corpus.\footnote{
    We ran question generation over
    GPT-4o, Llama 3.1 (8B), and Mistral summaries. As the resulting questions were highly similar, we did not include additional models.
}

\paragraph{Step 2: Clustering.}
We then cluster questions with the same semantic intent.
For instance, \emph{``Is the soundtrack effective?''} and \emph{``How does the music contribute to the film's atmosphere?''} are considered equivalent, as they ask for the same information. We select the question closest to the mean embedding of each cluster as its representative. These questions form the topics $T$.

\paragraph{Step 3: Question-Answering and Claim Decomposition.}
For each \texttt{\small (original document, question)} pair, we first obtain a \emph{reference answer} using a QA-model.
We then decompose each answer into a set of atomic claims $A_t$ (see~\cref{fig:overview} for an example).
These claims support the answerability calculation (described next), and a fine-grained analysis of summary similarity and consistency through claim entailment patterns (\cref{sec:results-model-model-similarity}).

\paragraph{Step 4: Answerability Estimation.}
We measure how well a summary answers a question by the fraction of reference answer claims it entails.
This naturally accounts for questions that are only partly answerable with a given summary.
Formally, let $A_t$ be the set of answer claims for a given question.
The answerability score is then calculated as:
\begin{equation}\label{eqn:answerability}
f(t,s) = \frac{1}{|A_t|} \sum\limits_{a \in A_t} e(a, s),
\end{equation}
where $e: A \times S \rightarrow \{0,1\}$ is a natural language inference (NLI) model that determines if claim $a$ is entailed (1) or not entailed (0) by summary $s$.
This practice of claim-entailment is commonly used in similar settings such as fact checking~\cite{Kamoi:2023:EMNLP,Min:2023:EMNLP,Stacey:2024:EMNLP}.\footnote{Since longer answers tend to include more claims, answer length may affect salience scores. In practice, we observe a weak negative relationship (see discussion in \cref{sec:appendix-salience-vs-answer-length}).}

\paragraph{Implementation.}
For question generation, we found it necessary to use a strong model (i.e., \gpt).
For clustering, following~\citet{Lam:2024:CHI}, we represent questions using sentence embeddings~\cite{Reimers:2019:EMNLP}, followed by a dimensionality reduction and density-based clustering with \textsc{hdbscan}~\cite{McInnes:2017:JOSS} which requires minimal parameter tuning and does not presuppose a fixed number of clusters.\footnote{
    We use \texttt{all-mpnet-base-v2} for sentence embeddings, \textsc{umap} for dimensionality reduction, and cluster with \textsc{hdbscan} (leaf clustering, min size = 15, defaults: $\epsilon=0$, $\alpha=1$).
}
After an initial round of clustering, we found several overlapping clusters which were merged manually.
For question-answering and answer-claim splitting, we use Llama 3.1 8B
(see \cref{sec:appendix-prompts} for the prompts).
For claim entailment,
we use the efficient \texttt{\small MiniCheck}~\cite{Tang:2024:EMNLP}.

\section{Experimental Settings}
\label{sec:experimental-settings}

\begin{table}[t]
\small
\centering
\setlength{\tabcolsep}{4pt}
\begin{tabular}{lrrrr}
\toprule
\textbf{Statistic} & \textbf{RCT} & \textbf{CL} & \textbf{Astro} & \textbf{QMSum} \\
\midrule
Documents & 200 & 185 & 106 & 90 \\
Words/doc & 290 & 459 & 703 & 10,837 \\
\midrule
Questions & 21 & 14 & 13 & 10 \\
Answered/doc & 84.1\% & 86.2\% & 96.5\% & 91.9\% \\
Words/answer & 30.9 & 53.0 & 70.3 & 161.5 \\
Claims/answer & 6.5 & 11.4 & 12.4 & 29.6 \\
Claims (total) & 23,124 & 25,353 & 16,430 & 24,459 \\
\bottomrule
\end{tabular}
\caption{Dataset overview. Number of words is calculated as whitespace-separated tokens.}
\label{tab:dataset-statistics}
\end{table}

\paragraph{Datasets.}
We analyze LLM salience across several technical and scientific domains using four datasets (\cref{tab:dataset-statistics}).
We designed slightly unconventional summarization tasks because of their limited ``oracle'' summaries in common LLM training data.
This allows us to analyze how LLMs handle texts without strong priors, and how salience judgments vary across genres and discourse types (technical writing, academic discourse, and dialogue).

\paragraph{(1) Randomized Controlled Trials (RCT).}
We draw a random sample of 200 abstracts of RCTs published Jan--Apr 2024 from PubMed.
These documents follow established conventions to describe the conduct and outcomes of clinical studies.
The task is to further summarize the abstracts.

\paragraph{(2) Computation and Language (CL).} The second task is to summarize the \emph{related work} sections of NLP/CL papers published on arXiv.
Although CL paper summarization is common, summarizing the related work section itself is not.
We convert raw LaTeX sources to Markdown and only consider documents up to 2,000 tokens to fit the context window of smaller models.
A random sample of 185 documents published in October 2024 is drawn.

\paragraph{(3) Astrophysics (Astro).} The third dataset contains \emph{discussion} sections of astrophysics papers published on arXiv.
These documents interpret key results of theoretical and empirical astrophysics research.
Similar to the CL portion, summarizing only the discussion sections is uncommon.
A random sample of 106 documents is drawn, with pre-processing analogous to CL.

\paragraph{(4) Meetings (QMSum).} Lastly, we consider meeting transcript summarization.
We randomly sample 90 documents balanced across three domains from QMSum~\cite{Zhong:2021:NAACL}: product design, research and political discussions.
We format transcripts as \texttt{\small [Speaker]: [Utterance]} turns, separated by newlines.
We only experiment with long-context models ($\ge 32\text{k}$ tokens) on this dataset.

\begin{figure*}[t]
\includegraphics[width=\textwidth]{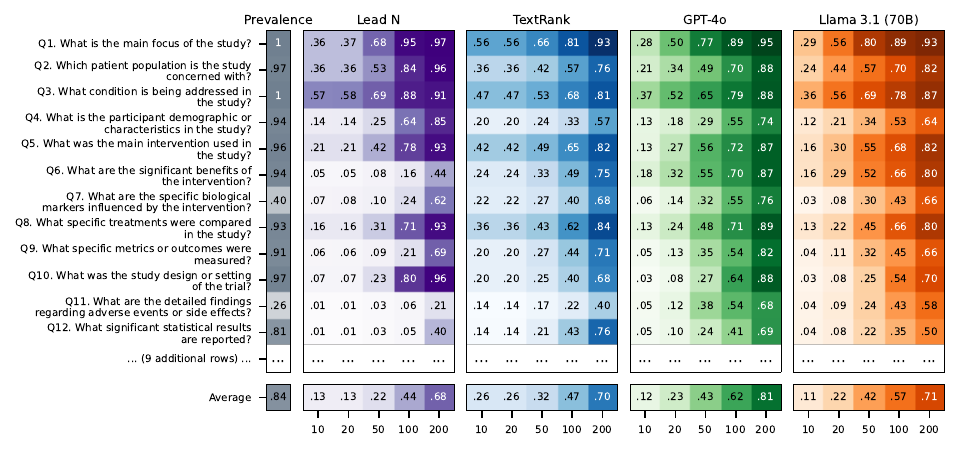}
\caption{Corpus-level content salience map for \emph{RCT} summaries by four methods (continued in~\cref{fig:salience-pubmed-full}).}
\label{fig:salience-pubmed}
\vspace{0.4em}
\end{figure*}

\paragraph{Summarization Models.}
We experiment with 13 LLMs of different scales:
\textbf{OLMo}~(7B; 02/24, 07/24; \citealp{Groeneveld:2024:ACL}), \textbf{Mistral}~(7B; v0.3; \citealp{Jiang:2023:arXiv}), \textbf{Mixtral}~(8x7B; v0.1, \citealp{Jiang:2024:arXiv}), \textbf{Llama 2}~(7B, 13B, 70B; \citealp{Touvron:2023:arXiv}), \textbf{Llama~3}~(8B, 70B), and \textbf{Llama~3.1}~(8B, 70B; \citealp{Grattafiori:2024:arXiv}). For API-based models, we use \textbf{\gptmini}~(07/24) and \textbf{\gpt}~(08/24; \citealp{OpenAI:2024:arXiv}).
We also include 3 baselines to contextualize results: \textbf{Lead-N}, \textbf{Random} and \textbf{TextRank}~\cite{Mihalcea:2004:EMNLP}, all adjusted to meet summary length budgets. To assess
consistency
across multiple rounds of decoding, we generate 5 summaries per document and target length with temperature $\tau = 0.3$.
We use a zero-shot summarization prompt (\cref{sec:appendix-prompts}).

Before analyzing salience in these models, we validate two key assumptions: \emph{(i)} generated summaries should approximately meet the target length, and \emph{(ii)} longer summaries should expand on shorter ones (``incremental consistency''). Additionally, we analyze how greater $\tau$ affect those criteria.
Our analysis confirms that models largely meet above criteria, with newer and bigger models showing better length control.
Higher $\tau$ results in stable \emph{average} summary length at the corpus level, but greater length variance at the document level (up to 10\% difference), along with a slight decline in incremental consistency (details in \cref{sec:appendix-length-analysis}).

\section{Observed Salience}
\subsection{RQ1: What notion of salience have LLMs learned in different domains?}
\label{sec:results-salience}
To understand how LLMs prioritize different information, we consider average question answerability as a proxy for salience. We show the results for the \emph{RCT} dataset as a representative example in \cref{fig:salience-pubmed}, and include other datasets in \cref{sec:appendix-salience}.

\textbf{Models prioritize information hierarchically.}
We observe a clear hierarchy in how information is prioritized across summary lengths.
For example, fundamental aspects such as the focus of a study (\emph{Q1}), and the condition being treated (\emph{Q3}) consistently achieve higher scores, even at 10-word summaries.
In contrast, more specific and technical information like the study design (\emph{Q10}) and the statistical significance of results (\emph{Q12}) are primarily discussed in longer summaries ($\ge 100$ words).

\textbf{Information frequency is not in itself predictive of salience.}
When we consider how frequently a question is answered by documents in the corpus (leftmost column of \cref{fig:salience-pubmed}), we find that even relatively rare questions such as biological markers and adverse effects (\emph{Q7}/\emph{11}, prevalence 40\%/26\%) maintain a consistent representation in summaries.
This suggests that LLMs do not simply prioritize information based on its frequency in a genre.

\textbf{Summaries progressively get more detailed, and information density differs across models.}
As expected, longer summaries consistently include more information as shown by the higher average answerability (bottom row in \cref{fig:salience-pubmed}).
However, the absolute scores differ across models.
GPT-4o has a notably higher answerability score than Llama 3.1, particularly at longer summaries (0.81 vs. 0.71 at the 200-word length).
Given that both models generate summaries of similar lengths (cf. \cref{fig:length-deviation}), this suggests that GPT-4o conveys information more efficiently.

\begin{figure*}[t]
\includegraphics[width=\textwidth]{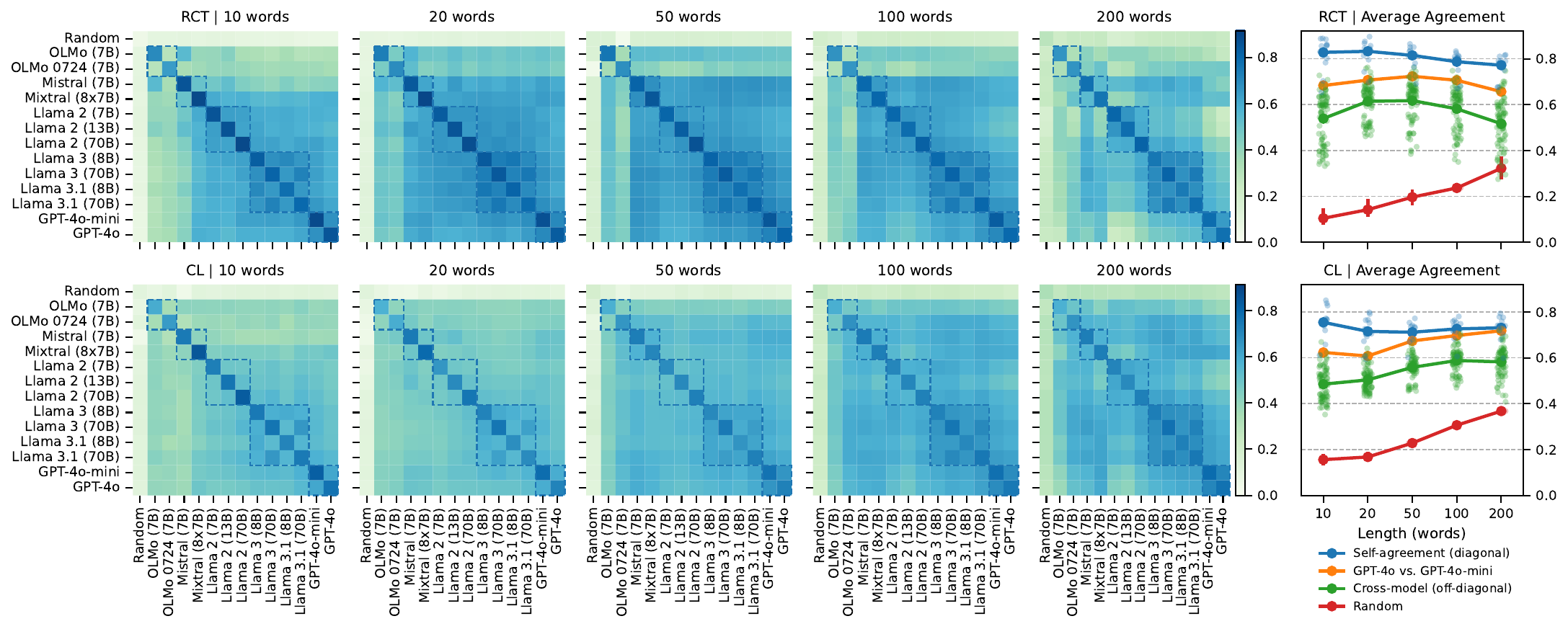}
\caption{
    Do LLMs share a similar notion of salience?
    Heatmaps show agreement of content-selection at the atomic-claim level (Krippendorff's $\alpha$).
    Dashed bounding boxes indicate models of the same family.
    The diagonal shows self-agreement over multiple generations. Top row: \emph{RCT}, Bottom row: \emph{CL}.
}
\label{fig:agg-pubmed-cl}
\vspace{-0.5em}
\end{figure*}

\subsection{RQ2: Do LLMs of different families and sizes have a similar notion of salience?}
\label{sec:results-model-model-similarity}
We want to understand to what extent different models (e.g., families, scales) have a shared notion of information salience in a given domain.
We define a fine-grained similarity metric that compares models' content-selection decisions.

Intuitively, two models are more similar if their summaries include the same answer claims.
More formally, for each summary length $l$, we compile all atomic claims derived from question-answers along with their entailment labels (cf. \cref{sec:method-questions}). These form a binary vector $\mathbf{v}_{M,l}$ indicating which claims model $M$ includes in its summaries.
We then measure agreement between two models using Krippendorff's alpha: $\alpha(\mathbf{v}_{M_1,l}, \mathbf{v}_{M_2,l})$.
This claim-level agreement metric is stricter than comparing aggregate answerability scores, as it requires models to consistently include or exclude the same claims at each summary length.\footnote{In contrast, similar answerability scores can result from selecting a similar \emph{number} of claims.}
\cref{fig:agg-pubmed-cl} shows the model-model agreement for the \emph{RCT} and \emph{CL} datasets.

\textbf{High agreement across multiple runs suggests models apply salience notion consistently.}
The diagonal in~\cref{fig:agg-pubmed-cl} shows the average pairwise agreement across 5 model runs.
Overall, self-agreement is the highest for \emph{RCT} ($\approx .80$), while it is slightly lower for \emph{CL}, \emph{Astro} and \emph{QMSum} ($\approx .75$).
We observe a slight decline in self-agreement as the summary length increases.
We hypothesize that each document has a tail of medium- to low-salient topics which may or may not be included as the length budget gives more ``freedom'' to the models.

In sum, high self-agreement suggests that models apply salience consistently, which is beneficial for downstream users who depend on predictable summarization behavior. Additionally, this result serves as a validation of our method and enables the following cross-model analyses which would be meaningless without high self-agreement.

\textbf{Models of the same family or size do \emph{not consistently} have a higher agreement than any other model.}
We next inspect the off-diagonal agreements, comparing one model family with another model family.
Overall, we find that within-family agreement is not consistently higher than cross-family agreement.
While there are isolated cases of a higher within-family agreement (e.g., Llama 3.1 and GPT-4o on \emph{RCT}), this trend cannot be confirmed for all families and datasets.

\begin{figure}[t]
\includegraphics[width=\linewidth]{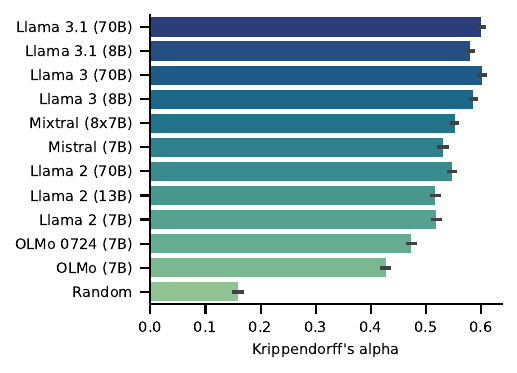}
\caption{Agreement with GPT-4o-mini, averaged over all datasets and summary lengths.}
\label{fig:agg-gpt4}
\end{figure}

\textbf{Agreement by summary length and with GPT-4o-mini.}
We observe that certain summary-lengths have higher agreement than others, though the peak is different for each dataset (e.g., agreement on \emph{RCT} is highest for 50 word summaries, whereas on \emph{CL} it peaks at 100 words).
There could be a ``natural'' summary length for each dataset where model more easily agree.
Lastly, we find that more recent and bigger models agree better with GPT-4o-mini which suggests a clear scaling effect and that open-weights models are getting closer in capabilities to large proprietary models (\cref{fig:agg-gpt4}).

\section{Perceived Salience and Alignment}
In addition to the \emph{observational salience} analysis,
we elicit \emph{perceived salience} by having humans and models directly rate the salience of each question.
This study has two purposes: (1) to understand whether model behavior aligns with human expectations, and (2) to see if the summarization behavior of LLMs can be approximated by direct prompting.

\subsection{Setup}
\paragraph{Human salience annotation.}
We recruited 18 experts across the four domains through our network (3 for \emph{RCT}, and 5 each for \emph{Astro, CL, QMSum}).\footnote{
    Trained physicians (\emph{RCT}), graduate students/faculty (\emph{Astro}), and graduate students (\emph{CL, QMSum}) based in US/Europe.
}
Experts rated the relative salience of each question on a 5-point Likert scale (ranging from 1: least important, to 5: most important).
Annotators were asked to motivate their rating through a brief rationale to encourage thoughtful judgments and to allow post-hoc analysis of their decision-making process.
To establish a shared understanding between annotators of what content a question may elicit, each question is accompanied by an example answer from a randomly drawn document in the domain.
To ensure high annotation quality, we conducted two pilot rounds with four annotators to refine our annotation guidelines (see \cref{sec:appendix-annotation-guidelines}).

Importantly, the human annotations cannot be regarded as a gold standard for salience. The ratings represent how humans \emph{perceive} question salience, which may not be reflective of how humans actually write summaries. As an initial step toward analyzing human salience through summarization behavior, we explore the application of our framework to human-written summaries in~\cref{sec:appendix-human-salience-news}.

\paragraph{Model-based salience ratings (LLM-perceived).}
We prompt LLMs to directly rate question salience.
The prompt includes the question list for a given domain and instructions that closely mirror the human annotation guidelines to allow for direct comparison (i.e., 5-point Likert scale and rationales).
Each model is prompted 5 times with a shuffled question list to mitigate position bias and to quantify consistency.
See \cref{sec:appendix-prompts} for the full prompt.

\paragraph{Analysis method.}
We use Spearman's rank correlation coefficient ($\rho$) to quantify alignment between three measures: human-perceived salience, LLM-perceived salience (both 5-point Likert), and LLM-observed salience (continuous $[0,1]$).\footnote{We take observed salience scores at the 200-words summary length which correlated on average most strongly with human salience. Other scores are explored in~\cref{sec:appendix-salience-score-ablation}.}
For groups with multiple ratings, we report averaged pairwise correlation and test for statistical significance with the harmonic-mean p-value~\cite{Wilson:2019:PNAS}.

\paragraph{Human correlation.}
Inter-human correlation varies by domain, with meeting summarization (QMSum, $\rho = 0.60$) and RCT abstracts ($\rho = 0.46$) showing a moderate to strong correlation (\cref{tab:annotator-agreement}).
These domains presumably have established conventions about summary content.
In contrast, correlation is weak for summarization of related work (CL, $\rho = 0.26$) and discussion sections (Astro, $\rho = 0.16$).
Documents in these domains may vary significantly in the type of content they present (i.e., certain questions may be more relevant to theoretical vs. empirical papers).
While our annotation protocol aims to control for this aspect through the example answers by question, there remains annotator subjectivity related to their personal interests.

\begin{table}[t]
\sisetup{
detect-weight,
mode=text,
table-format=-1.2,
table-space-text-post={$^{**}$}
}
\small
\centering
\begin{tabular}{lccSc}
\toprule
\bfseries Dataset & \bfseries Questions & \bfseries Raters & \B {Spearman} & \bfseries Std. \\
\midrule
QMsum & 10 & 5 & 0.60$^{*}$ & 0.18 \\
RCT & 21 & 3 & 0.46$^{*}$ & 0.06 \\
CL & 14 & 5 & 0.26$^{**}$ & 0.29 \\
Astro & 13 & 5 & 0.16 & 0.44 \\
\bottomrule
\end{tabular}
\caption{Inter-annotator correlation for question salience rating. Significance: $^{*}$ ($p < 0.05$) and $^{**}$ ($p < 0.01$).}
\label{tab:annotator-agreement}
\end{table}

\subsection{Results}
\label{sec:results-introspection}
To understand if LLMs can reliably rate question salience, we study three conditions.
First, as a reference point, we measure consistency of the observational and perceived salience measures estimated over 5 model runs (LLM-observed, LLM-perceived).
Second, we study the correlation of LLM-perceived and LLM-observed to measure if models' explicit ratings align with their summarization behavior (RQ3).
Third we correlate LLM-derived salience in human perceived salience (RQ4).
We report results for the three conditions in \cref{tab:results-rater-agreement} and provide qualitative examples in \cref{tab:results-examples}.

\begin{table*}[t]
\sisetup{
detect-weight,
mode=text,
table-format=-1.2,
table-space-text-post={$^{**}$}
}
\small
\centering
\setlength{\tabcolsep}{4pt}
\begin{tabular}{lSSSSSSSS}
\toprule
\B {Measure} & \B {Random} & \B {OLMo} & \B {Mixtral} & \B Llama$^{3.1}_{8b}$ & \B Llama$^{3.1}_{70b}$ & \B {4o-mini} & \B {4o} & \B {Average} \\
\midrule
\multicolumn{9}{c}{\texttt{Consistency of Salience Estimates}}\\
\emph{LLM-perceived} & -0.05 & 0.20$^{*}$ & 0.54$^{**}$ & 0.37$^{*}$ & 0.71$^{**}$ & 0.73$^{**}$ & \B 0.76$^{**}$ & 0.47$^{**}$ \\
\emph{LLM-observed} & 0.92$^{**}$ & \B 0.99$^{**}$ & \B 0.99$^{**}$ & 0.98$^{**}$ & \B 0.99$^{**}$ & 0.98$^{**}$ & 0.98$^{**}$ & 0.97$^{**}$ \\\addlinespace
\multicolumn{9}{c}{\texttt{Correlation of Salience Estimates}}\\
\emph{LLM-perceived vs. -observed} & 0.03 & 0.12 & 0.37$^{*}$ & 0.36$^{**}$ & 0.47$^{*}$ & \B 0.56$^{**}$ & 0.50$^{*}$ & 0.35$^{*}$ \\\addlinespace
\multicolumn{9}{c}{\texttt{Correlation of Model and Human Salience}}\\
\emph{LLM-perceived vs. Human} & 0.07 & 0.16 & 0.41$^{*}$ & 0.31$^{*}$ & 0.46$^{**}$ & 0.51$^{**}$ & \B 0.53$^{**}$ & 0.35$^{**}$ \\
\emph{LLM-observed vs. Human} & 0.20 & 0.25 & 0.33$^{*}$ & 0.35$^{*}$ & \B 0.36$^{*}$ & 0.34$^{*}$ & 0.25 & 0.30$^{*}$ \\
\bottomrule
\end{tabular}
\caption{Spearman rank correlations between salience estimates, averaged across datasets. Per-dataset values in \cref{tab:results-rater-agreement-full}. Significance: $^{*}$ ($p < 0.05$) and $^{**}$ ($p < 0.01$); row-wise maxima are bolded.}
\label{tab:results-rater-agreement}
\end{table*}

\begin{table*}[t]
    \fontsize{8.5pt}{8.5pt}\selectfont
    \begin{tabular}{@{}p{0.17\textwidth}rp{0.23\textwidth}p{0.44\textwidth}@{}}
    \toprule
        \textbf{Question} &
        $L_o$/$L_p$/$H$ &
        \faRobot\ \textbf{Rationale LLM-perceived} &
        \faUsers\ \textbf{Rationales Human}\\
    \midrule
    \textbf{(\#1, RCT \faUserMd)}
        What significant statistical results are reported? & 2.5/4.0/5.0 &
        Statistical results are essential for assessing the validity and reliability of the findings. &
        (H\textsubscript{\{1,2,3\}} = 5) Main outcome that physicians are looking for; very important to be included as measure of success; without it reporting is lacking \\

    \textbf{(\#2, Astro \faStar)}
        What broader context or field does the study contribute to or address?
        & 4.6/3.4/2.4
        & Understanding the broader context helps situate the study within the larger scientific discourse.
        & (H\textsubscript{\{2,5\}} = 1) not relevant, experts would fill context; (H\textsubscript{1} = 3) inferable from another question; (H\textsubscript{3} = 4) context is important although not always addressed; (H\textsubscript{4} = 3) Situating findings within literature, existing models or theories is important\\

    \textbf{(\#3, RCT \faUserMd)}
        What was the main intervention used in the study? & 4.2/4.8/5.0 &
        The main intervention is central to understanding the study's focus and findings. &
        (H\textsubscript{\{1,2,3\}} = 5) determines whether study influences decision making; part of PICO therefore fundamental; essential for communicating purpose/focus of the trial \\
    \bottomrule
    \end{tabular}
    \caption{Example questions, salience scores by LLM-observed ($L_o$, rescaled to 1-5), LLM-perceived ($L_p$), humans ($H$) and summarized rationales. Additional examples in \cref{tab:results-examples-part2}.}
    \label{tab:results-examples}
\end{table*}

\paragraph{RQ3: When models introspect, does their perceived notion of salience align with their summarization behavior?}
LLMs have strong and consistent \emph{implicit} notions of salience, but they are unreliable when explicating these preferences in rating tasks. We detail these observations below.

\textbf{Observational salience is highly stable.}
We find that observational question salience leads to highly stable scores for all models ($\rho \ge 0.98$).
This suggests that LLMs' underlying summarization process is highly deterministic despite the stochastic nature of language models.
Also, it suggests that our proposed approach is a reliable tool for analyzing model behavior.

\textbf{Models fail to have consistent perceived salience.}
We find that the consistency of direct salience ratings varies significantly for all models and datasets.
Generally, strong instruction-following models have more consistent perceived salience than weaker models (avg. $\rho$ ranges from 0.20 for OLMo to 0.76 for GPT-4o).
This finding mirrors recent results in the LLM-as-a-judge literature which demonstrated instability in ratings due to various factors including position bias~\cite{Wang:2024:ACL,Stureborg:2024:arXiv}.

\textbf{Perceived $\neq$ observed salience.}
Lastly, we find only a weak to moderate correlation between perceived and observed salience (highest: avg. $\rho = 0.56$ for GPT-4o-mini, lowest: $\rho = 0.12$ for OLMo).
Again, stronger instruction-following models show higher correlations, indicating a clear scaling effect.
This gap echoes broader findings where generative abilities may not reflect an underlying understanding in models~\cite{West:2024:ICLR}.

\paragraph{RQ4: To what extent does model salience align with human perceived salience?}
\label{sec:results-human-alignment}
We find that both LLM-salience estimates only show a weak to moderate correlation with human salience perception.
Direct rating for question salience correlates more than observed salience (highest LLM-perceived: avg. $\rho = 0.53$ for GPT-4o, highest LLM-observed: avg. $\rho = 0.36$ for Llama 3.1 70B).
Weak correlation between models and humans holds for all dataset, also those where humans agree more strongly among themselves (\cref{tab:results-rater-agreement-full}).

In sum, LLM users should carefully consider if a model is appropriate for their summarization task, or provide explicit signals about content priority through prompts or during model training.

\section{Related Work}
\paragraph{Evaluating and Interpreting Summarization.}
Recent work suggests that LLMs match or surpass human performance in news summarization~\cite{Zhang:2024:TACL}.
However, traditional evaluation protocols remain unreliable especially for LLM-generated summaries~\cite{Fabbri:2021:TACL,Goyal:2023:arXiv}.
This spurred interest in analyzing summarization model behavior.
Studies found biases towards content near the beginning/end of documents~\cite{Ravaut:2024:ACL,Laban:2024:EMNLP}.
Others analyze training dynamics of summarization models to identify when skills like content selection are learned~\cite{Goyal:2022:ACL}.
Extract-then-abstract pipelines~\cite{Gehrmann:2018:EMNLP,Li:2021:ws} aim for interpretable text summarization but this interpretability is limited to the document-level~\cite{Dhaini:2024:INLG}.
Our research complements prior work by providing a \emph{global interpretation} of what topics LLMs consider important through the lens of text summarization.

\paragraph{Explainable Topic Modeling.}
Our analysis method draws inspiration from the interpretable topic modeling literature.
While classical topic models such as LDA~\cite{Blei:2003:JMLR} have long been used to explain latent themes in text corpora, they are often difficult to interpret~\cite{Chang:2009:NeurIPS}.
Recent work showed that LLMs can effectively be used to generate natural language descriptions of latent themes in text mining, clustering and concept induction workflows~\cite{Pham:2024:NAACL,Zhong:2024:NeurIPS,Wang:2023:EMNLP,Lam:2024:CHI}.
Our framework uses LLMs to describe salient summary content in form of information-seeking QUDs.
The use of QUDs as a representation of information units was shown successful in a wide range of tasks~\cite{Newman:2023:EMNLP,Laban:2022:Findings,Trienes:2024:ACL,Wu:2023b:EMNLP}.
Finally, in the context of summarization, our work shares theoretical foundations with~\citet{Wu:2024:EMNLP}, who explore human curiosity through inquisitive QUDs. They observe that answering salient questions is a quality indicator for news summaries.

\section{Conclusion}
We propose an interpretable framework to systematically derive and analyze LLMs' notion of information salience, a previously latent concept that is nonetheless crucial for text synthesis applications.
Our work builds on two key ideas: using length-controlled summarization as a behavioral probe for content selection, and describing what is salient as the answerability of questions.
We found that LLMs have a highly consistent notion of salience which is largely compatible across models.
We further found that LLMs cannot directly rate the salience of questions, and that model salience weakly aligns with human perceptions.
Our work opens new directions to study how LLM salience emerges during training, and for diagnosing content selection challenges in text synthesis tasks.

\section*{Limitations}
We consider zero-shot prompting with temperature-based decoding to generate summaries. While these settings are common defaults for LLM users, it is conceivable that different prompting styles (e.g., chain-of-density) or decoding methods influence salience patterns. Future work should explore how these techniques affect salience, particularly in adjacent information-seeking tasks such as query-based summarization.

While our experiments cover diverse disciplines (medicine, astrophysics, computational linguistics, and meetings) and discourse types (structured writing, academic discourse, and dialogue), the texts are primarily technical. Since our framework is designed to be domain-agnostic, we believe it is an exciting direction for future work to explore less technical genres such as fiction~\cite{Kim:2024:COLM}.

Our user study assumed a uniform background and interests among participants, which is a simplification of practical applications. Additionally, the specialized nature of two tasks (i.e., summarizing related work and discussion sections) may have contributed to variability in responses, as even domain experts may not have strong priors on how these texts should be summarized. Future work could explore how differences in expertise and prior knowledge shape perceptions of salience.

\section*{Acknowledgments}
We thank Hsin-Pei Chen, Khawla Elhadri, Arya Farahi, Juan P. Farias, Cheng Han Hsieh, Sebastian Joseph, Ramez Kouzy, Michael Muzinich, Van Bach Nguyen, Juan Diego Rodriguez, Paul Torrey, Manya Wadhwa, Barry Wei, and Paul Youssef for their participation in the salience annotation study.
We also thank Dennis Aumiller and Philippe Laban for early feedback on this research.
This work was partially supported by the US National Institutes of Health (NIH) grant 1R01LM014600-01,
and the US National Science Foundation grants IIS-2107524, IIS-2145479, and Cooperative Agreement 2421782 and Simons Foundation MPS-AI-00010515 (NSF-Simons AI Institute for Cosmic Origins\footnote{CosmicAI, \url{https://www.cosmicai.org/}}).

\bibliography{bibliography}

\clearpage
\appendix
\section{Length-instruction Following}
\label{sec:appendix-length-analysis}
We analyze to what extent length-controlled summarization is a consistent probe for content selection.
Ideally, we expect the following behavior of summarization models: (1) the generated summary length matches approximately the target length, and (2) as we increase the length budget, summaries should provide all content of the shorter version in addition to expanding on it.
We define two measures for these desiderata.

\paragraph{Target length ratio (TLR).}
We quantify the length deviation of a generated summary ($s_{l}$) from the target word count ($l$) as follows:
\begin{equation}
\text{\textbf{TLR}}(s_{l}) = \frac{|s_{d,l}|}{l}.
\end{equation}
Where $|\cdot|$ is the summary length (whitespace separated tokens). A value of 1 indicates perfect length match, while values greater or smaller than 1 indicate over- or under-generation, respectively.

\paragraph{Incremental consistency (IC).}
Longer summaries should contain a proper superset of claims found in the adjacent shorter version.
Formally, for each document $d$ and topic $t$ recall that we have a set of atomic claims $A_t$ (\cref{sec:method-questions}).
We first identify the set of claims that are entailed at least once across any summary length:
\begin{equation*}
    A_{\text{entailed}}(d,t) = \{a \in A_t \mid \exists l \in L, e(a,s_{d,l}) = 1\},
\end{equation*}
where $e$ is an NLI model indicating whether claim $a$ is entailed by summary $s_{d,l}$ of length $l$.
Next, we determine if a claim is included consistently across increasing summary lengths (monotonicity condition).
\begin{equation*}
    e(a,s_{d,l_1}) \leq e(a,s_{d,l_2}) \; \forall l_1 < l_2
\end{equation*}
We then define the set of consistent claims where this condition holds:
\begin{equation*}
\begin{aligned}
    A_{\text{consistent}}(d,t) =\,&\{a \in A_{\text{entailed}}(d,t) \\
    \quad& \mid \text{\small monotonicity holds} \, \forall l \in L \}.
\end{aligned}
\end{equation*}
Finally, the overall incremental consistency for summaries of a corpus $D$ is given as the fraction of consistent claims:
\begin{equation}
\text{\textbf{IC}}(D) = \frac{
        \sum_{d \in D} \sum_{t \in T} |A_{\text{consistent}}(d,t)|
    }{
        \sum_{d \in D} \sum_{t \in T} |A_{\text{entailed}}(d,t)|
    }.
\end{equation}
This metric ranges from 0 to 1, where 1 indicates perfect monotonicity (longer summaries always include all information found in shorter ones).

\begin{figure}[t]
\includegraphics[width=\linewidth]{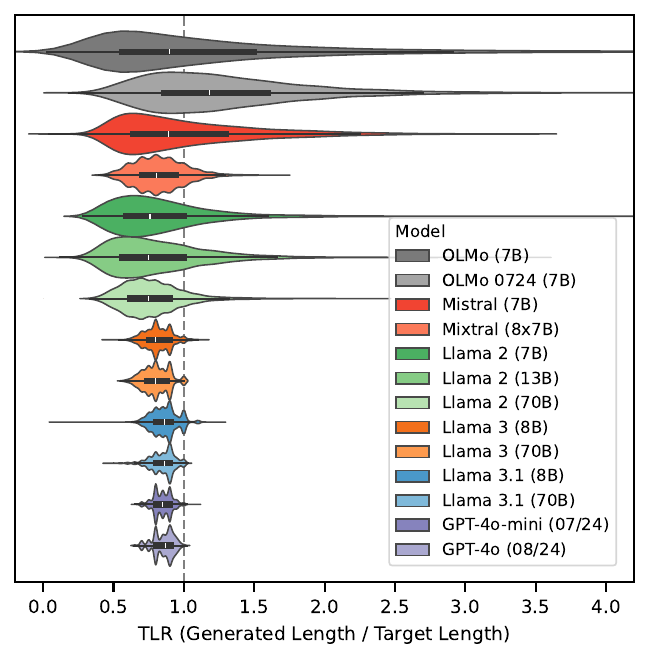}
\caption{Distribution of target length ratios over all generated summaries (aggregating lengths and datasets).}
\label{fig:length-deviation}
\end{figure}

\begin{figure*}[t]
\centering
\begin{subfigure}[b]{\textwidth}
\includegraphics[width=\textwidth]{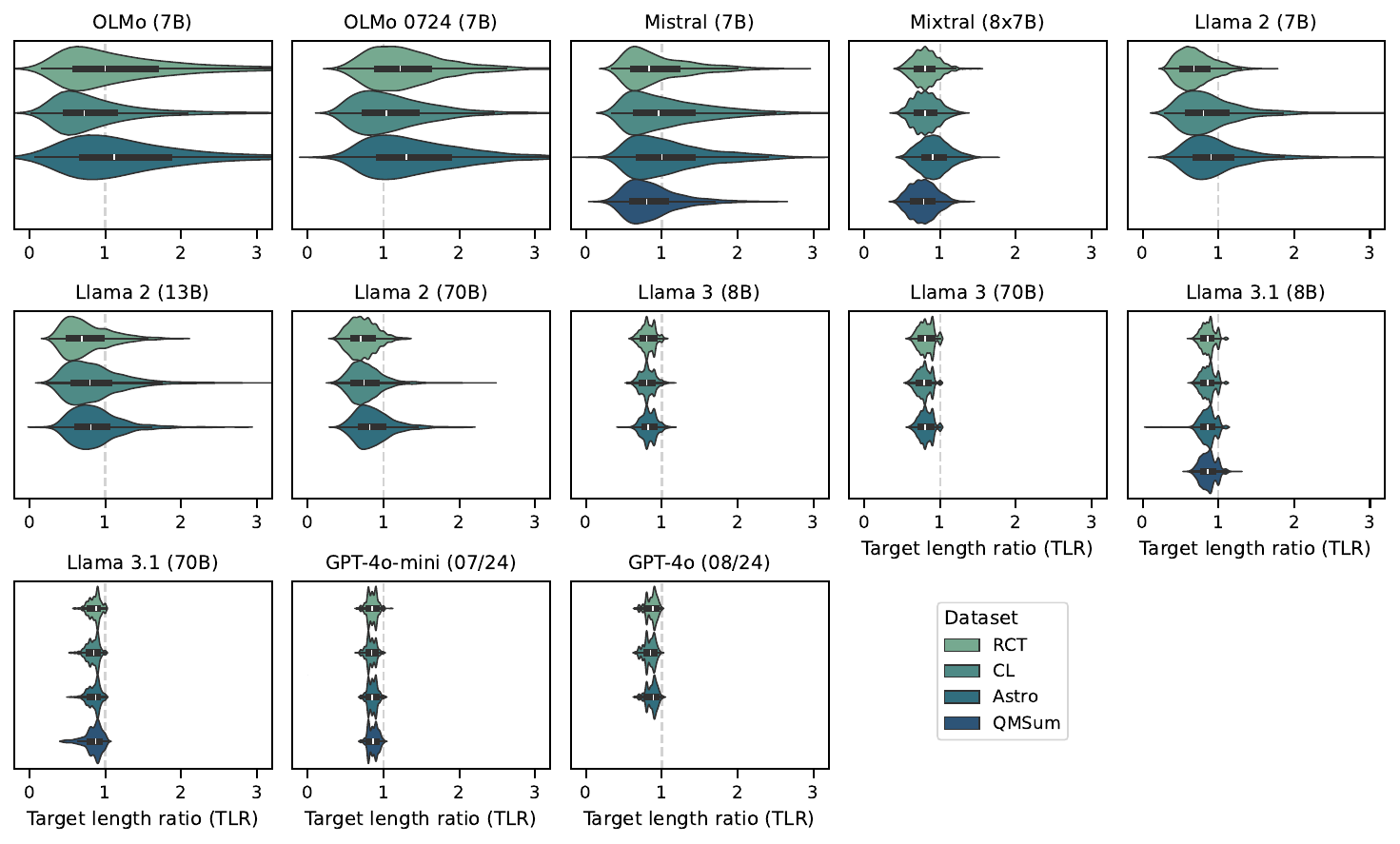}
\caption{Distribution of target length ratios over all generated summaries stratified by dataset.}
\label{fig:tlr-dataset}
\end{subfigure}
\vspace{0.5cm}
\begin{subfigure}[b]{\textwidth}
\includegraphics[width=\textwidth]{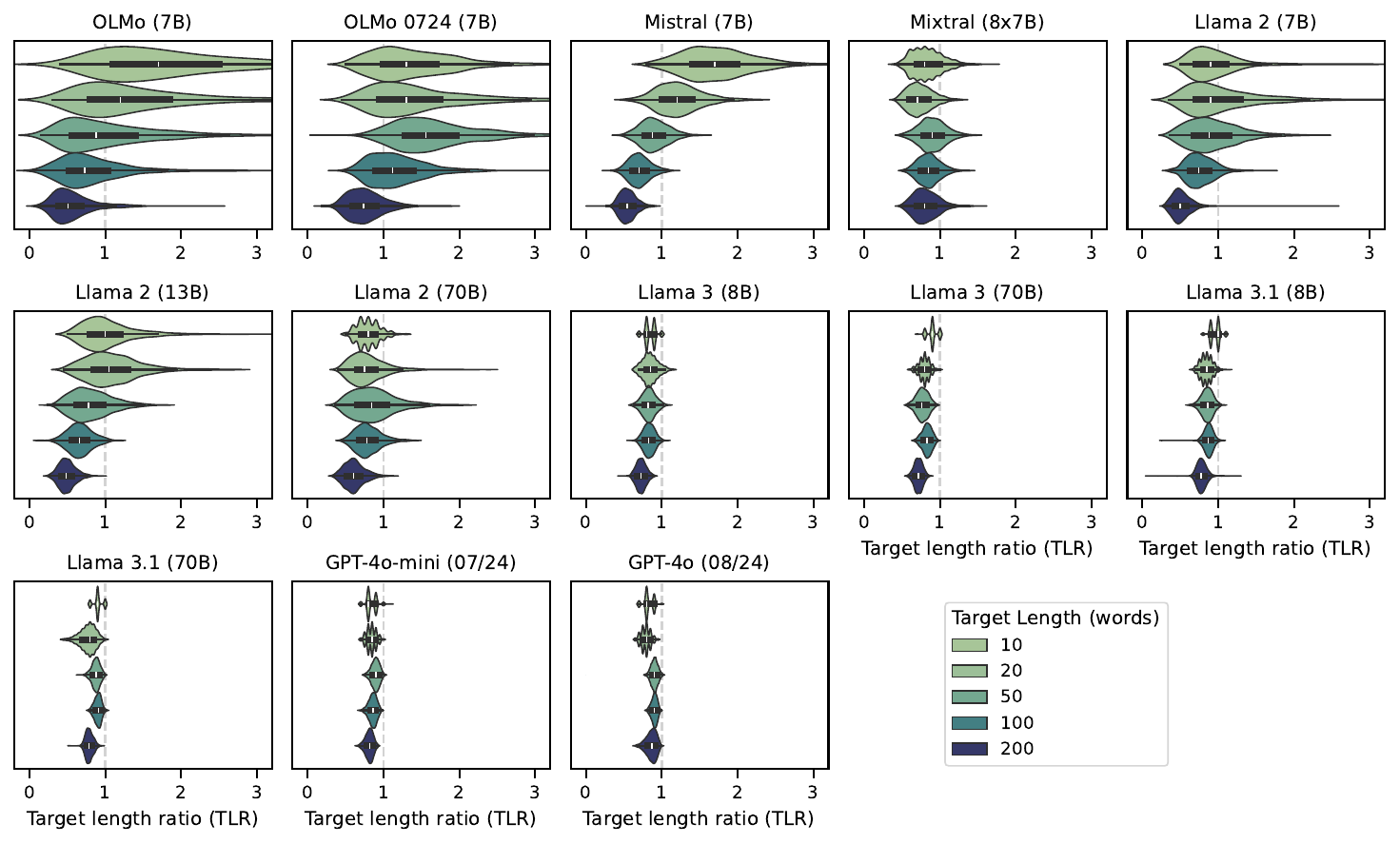}
\caption{Distribution of target length ratios over all generated summaries stratified by target summary length.}
\label{fig:tlr-length}
\end{subfigure}

\caption{Analysis of length-instruction following. The target length ration (TLR) indicates to what extent models match the provided length. A value of 1 indicates perfect length match, while values greater or smaller than 1 indicate over- or under-generation, respectively.}
\label{fig:tlr-stratified}
\end{figure*}

\paragraph{Do models meet the target length?}
We find that all models generally undershoot the length target (\cref{fig:length-deviation}).
However, more recent models match the target length more closely and consistently, showing a clear scaling effect.
The best performing models are Llama 3.1 and GPT-4o, while OLMo is unable to follow length-instructions, presumably because this was not part of the instruction tuning data.
Surprisingly, we do not find substantial differences across datasets.
This suggests that the ability of models to follow length-instructions is mostly invariant to the input document length, even if they are considerably long (e.g., meeting transcripts).
See \cref{fig:tlr-stratified} for an analysis of summary length stratified by dataset and target length.

\begin{figure}[t]
\includegraphics[width=\linewidth]{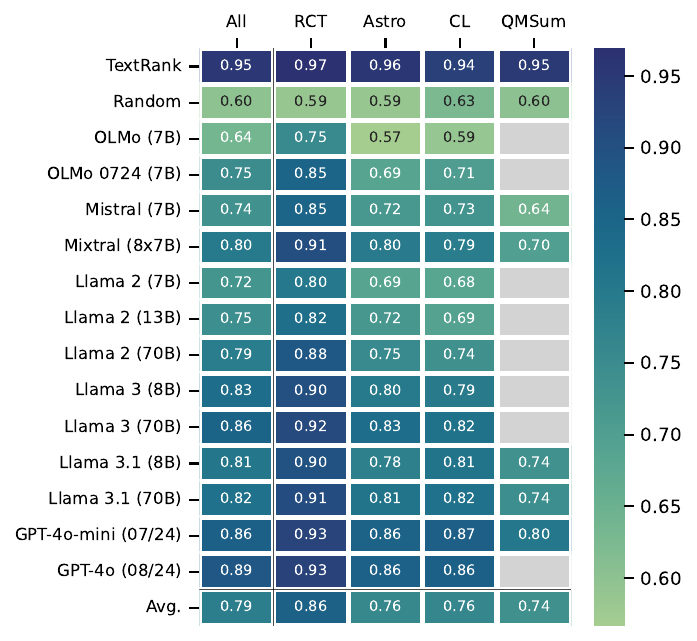}
\caption{Incremental consistency by model and dataset.}
\label{fig:incremental-consistency}
\end{figure}

\paragraph{How incrementally consistent are summaries?}
We report the average incremental consistency by dataset and model in \cref{fig:incremental-consistency}.
We observe that all models are substantially more consistent than the random summarization baseline.
Furthermore, incremental consistency decreases with more difficult datasets, likely because there is more freedom on what content to include in a summary.
Similar to the ability of following length instructions, we observe a scaling effect where stronger models have a higher incremental consistency.

\begin{figure}[t]
\includegraphics[width=\linewidth]{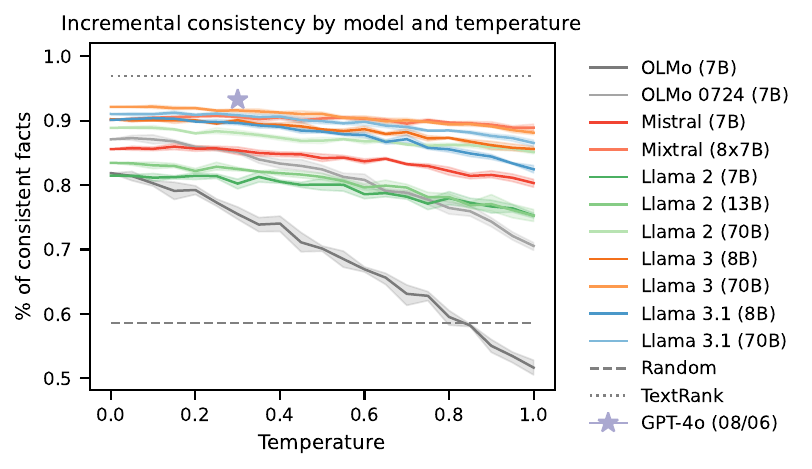}
\caption{Incremental consistency by temperature.}
\label{fig:temperature-ic}
\end{figure}

\begin{figure*}[t]
\includegraphics[width=\textwidth]{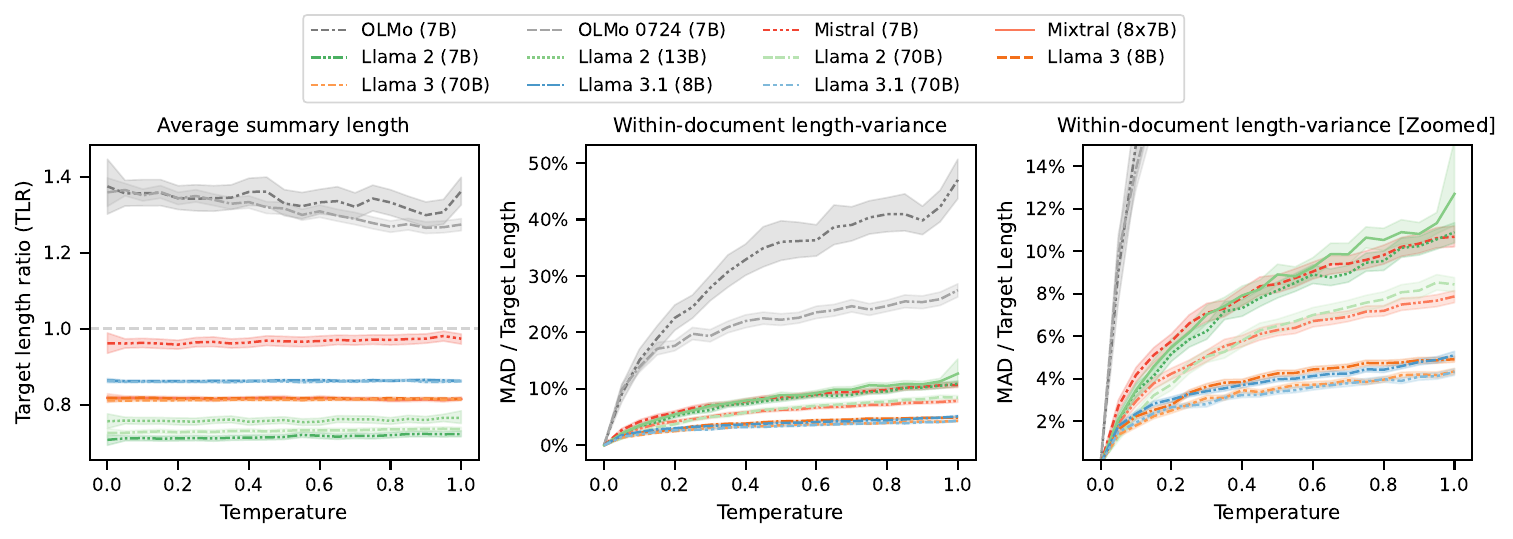}
\caption{Influence of temperature on generated summary length. \textbf{Left:} target length-ratio. \textbf{Center:} ``within-document length variance'' calculated as the mean deviation from the average summary length of 5 summaries for the same document (MAD). MAD is normalized to be comparable across length targets. \textbf{Right:} zoomed version.}
\label{fig:temperature-length}
\end{figure*}

\paragraph{Influence of temperature sampling.}
The main results in this paper are obtained with a temperature of $\tau = 0.3$.
To assess how temperature affects summary length and incremental consistency, we perform a temperature sweep on the RCT dataset for all open-weights models (20 settings in $[0,1]$).
Surprisingly, higher temperatures do not affect the \emph{average} summary length on a dataset-level, but lead to greater variance at the document level (up to 10\% length difference between generations, \cref{fig:temperature-length}).
Furthermore, higher temperatures lead to a slight decline in incremental consistency for all models that adequately follow length instructions (a drop of 1\% to 9\%, \cref{fig:temperature-ic}).

\paragraph{Summary.}
Overall, we find that strong models are able to follow length-instructions and that they consistently expand the summary content with increasing length budgets.
As our salience analysis assumes this behavior of models, it may be less reliable for weaker models (OLMo, Mistral, Llama~2).

\section{Salience Analysis}
\label{sec:appendix-salience}
\begin{figure*}[t]
\includegraphics[width=\textwidth]{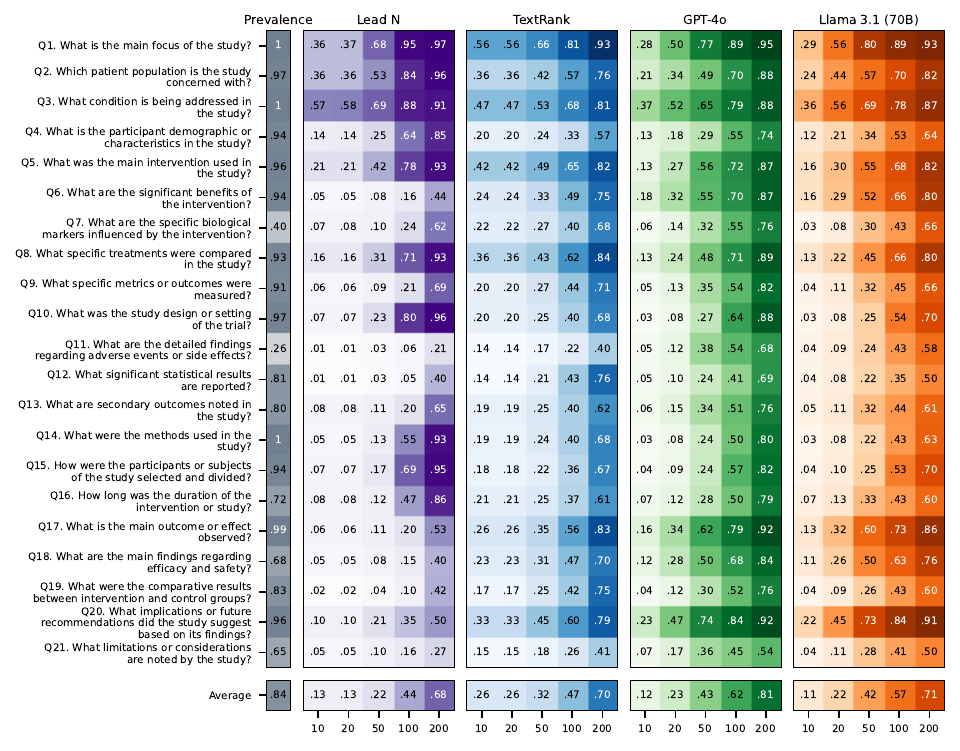}
\caption{Corpus-level content salience map for \emph{RCT} summaries by four methods.}
\label{fig:salience-pubmed-full}
\end{figure*}

\begin{figure*}[t]
\includegraphics[width=\textwidth]{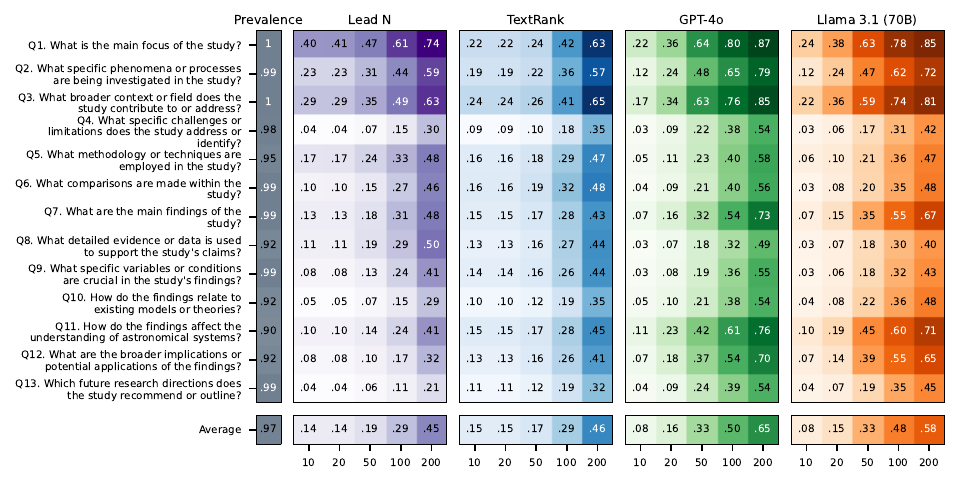}
\caption{Corpus-level content salience map for \emph{Astro} summaries by four methods.}
\label{fig:salience-astro-ph}
\end{figure*}

\begin{figure*}[t]
\includegraphics[width=\textwidth]{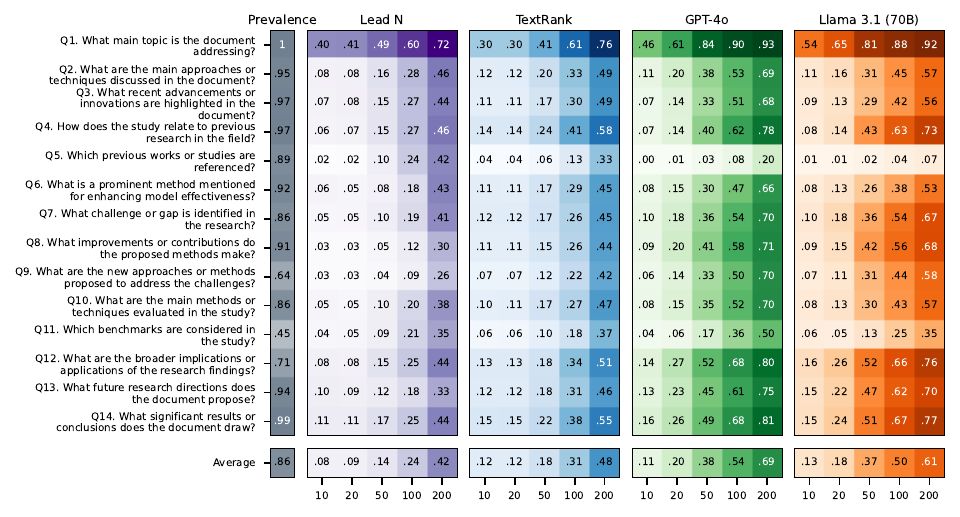}
\caption{Corpus-level content salience map for \emph{CL} summaries by four methods.}
\label{fig:salience-cs-cl}
\end{figure*}

\begin{figure*}[t]
\includegraphics[width=\textwidth]{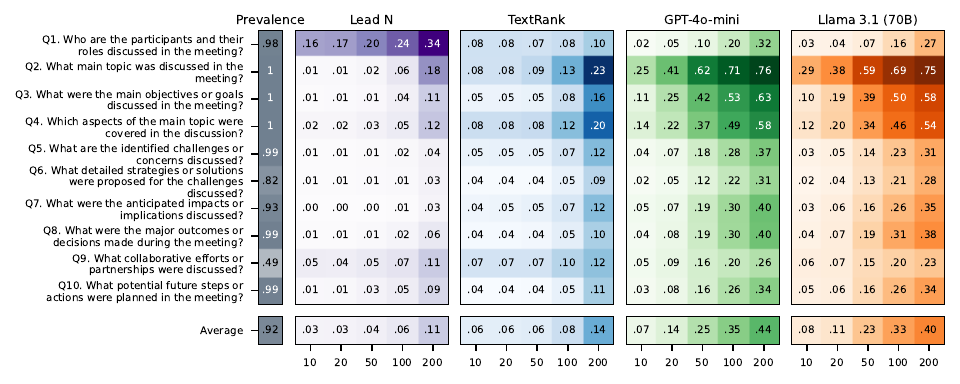}
\caption{Corpus-level content salience map for \emph{QMSum} summaries by four methods.}
\label{fig:salience-qmsum}
\end{figure*}

\begin{figure*}[t]
\includegraphics[width=\textwidth]{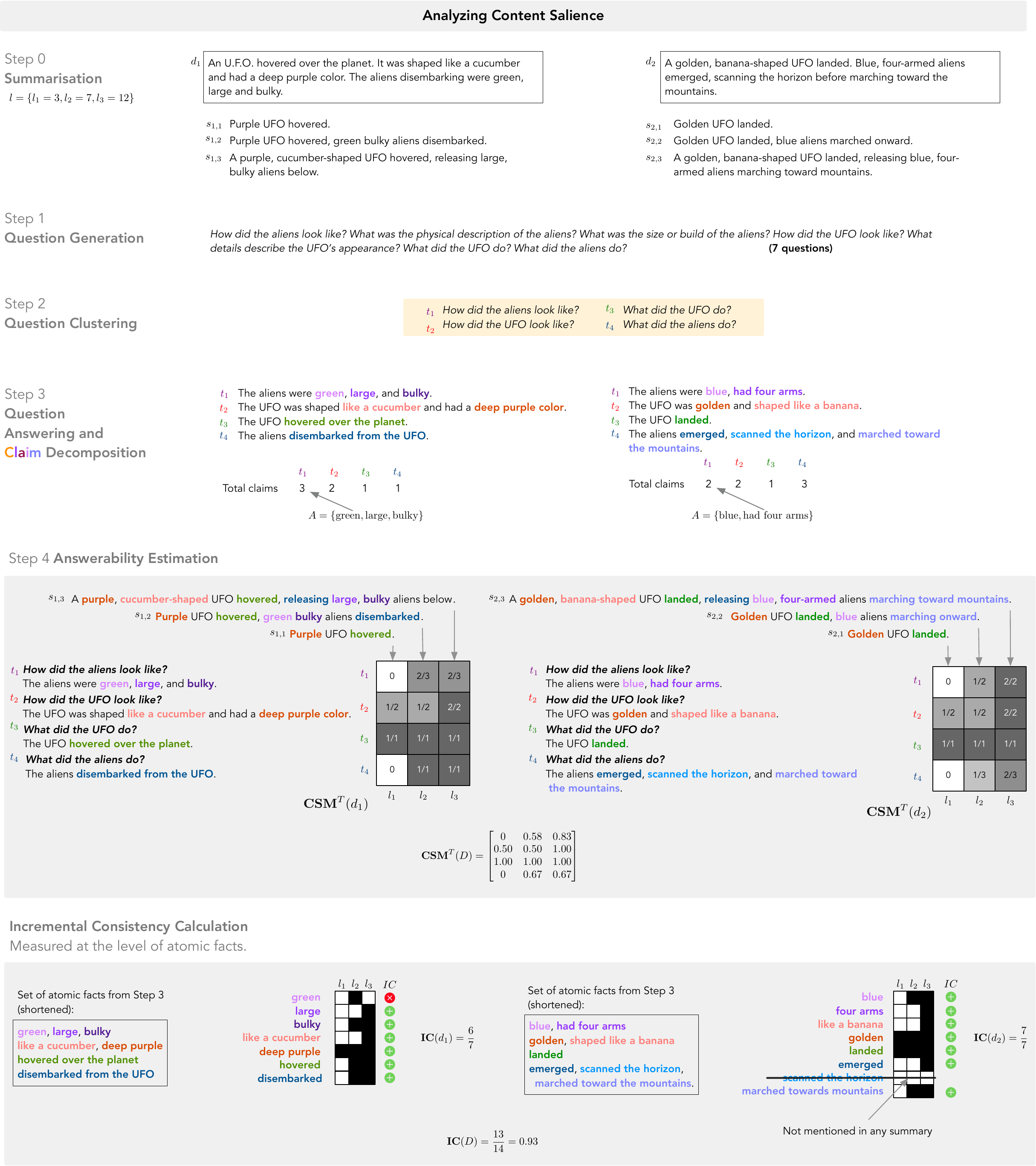}
\caption{Fully worked example of the question-based content analysis. Two documents in a fictional domain are each summarized at three lengths. Afterwards Steps 1 -- 4 are analogous to \cref{sec:method-questions}. Summary claims are color-coded.}
\label{fig:worked-example}
\end{figure*}

The corpus-level salience analysis for PubMed, Astro, CL, and QMSum is given in \cref{fig:salience-pubmed-full}, \cref{fig:salience-astro-ph}, \cref{fig:salience-cs-cl}, and \cref{fig:salience-qmsum}, respectively. We also provide a fully-worked example of the content salience analysis in \cref{fig:worked-example}.

\section{Ablation: Salience Score}
\label{sec:appendix-salience-score-ablation}
We analyze how different salience scores derived from the CSM correlate with human salience.
Recall that the $\text{CSM}(D)_{t,l}$ tracks the average answerability of question $t \in T$ at summary length $l \in L = \{10,20,50,100,200\}$.
We take raw salience scores at each summary length. Additionally, we calculate several question-wise aggregate scores.
Intuitively, questions which are more answerable at shorter summaries score higher under the aggregated scheme.
Formally, we aggregate scores as follows:
\begin{equation*}
\text{CSM}_{\text{agg}}(D)_{t} = \frac{\sum_{l \in L} w_l \cdot \text{CSM}(D)_{t,l}}{\sum_{l \in L} w_l},
\end{equation*}
where $w_l$ is a weighting term. We experiment with three weighting functions: uniform ($w_l = 1$), reciprocal length ($w_l = 1/l$), and logarithmic decay ($w_l = 1/\log(1 + l)$).

\begin{figure}[t]
\includegraphics[width=\linewidth]{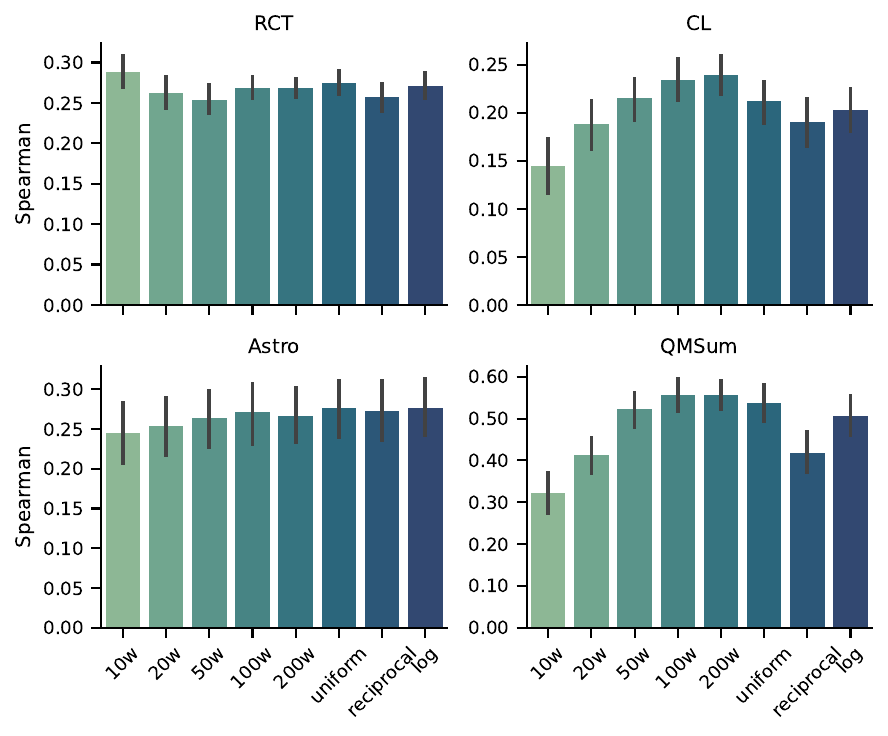}
\caption{Correlation of different salience scores with human salience. Here we aggregate over all LLMs which showed similar trends.}
\label{fig:salience-score-correlation}
\end{figure}

\cref{fig:salience-score-correlation} shows the Spearman rank correlation coefficient ($\rho$) with human salience for each salience score.
Overall, we find that all salience scores correlate similarly with human salience ratings on \emph{RCT} and \emph{Astro}, while the 200 words salience score correlates most strongly on \emph{CL} and \emph{QMSum}.

\section{Ablation: Effect of Answer Length on Question Salience}
\label{sec:appendix-salience-vs-answer-length}
We calculate question salience as the fraction of answer claims entailed by the summary (\cref{eqn:answerability}). Naturally, some questions can be answered succinctly (e.g., \emph{``What is the goal of the study?''}) while others require more elaboration (e.g., \emph{``What were the detailed findings?''}).
This raises the question how answer length influences salience. To better understand this relationship, we compute the Spearman rank correlation between answer length (measured in whitespace-separated tokens) and question salience.
\cref{tab:salience-vs-answer-length} reports results for summaries generated by Llama 3.1 (70B), with similar trends for other models.

\begin{table}[t]
\small
\centering
\sisetup{
detect-weight,
mode=text,
table-format=-1.2,
table-space-text-post={$^{**}$}
}
\begin{tabular}{lSSSS}
\toprule
\B {Summary} & \B {RCT} & \B {Astro} & \B {CL} & \B {QMsum} \\
\midrule
10 words & -0.03 & -0.11$^{**}$ & -0.07$^{**}$ & -0.22$^{**}$ \\
20 words & -0.10$^{**}$ & -0.17$^{**}$ & -0.10$^{**}$ & -0.31$^{**}$ \\
50 words & -0.18$^{**}$ & -0.21$^{**}$ & -0.19$^{**}$ & -0.36$^{**}$ \\
100 words & -0.26$^{**}$ & -0.28$^{**}$ & -0.22$^{**}$ & -0.41$^{**}$ \\
200 words & -0.31$^{**}$ & -0.31$^{**}$ & -0.26$^{**}$ & -0.45$^{**}$ \\
\bottomrule
\end{tabular}
\vspace{-0.2em}
\caption{Spearman rank correlation between \emph{answer length} and \emph{question salience} for summaries generated by Llama 3.1 (70B). Significance: $^{**}$ ($p < 0.01$).}
\label{tab:salience-vs-answer-length}
\vspace{-0.2em}
\end{table}

We observe a weak negative correlation between answer length and question salience for shorter summaries (10--20 words), and a weak-to-moderate negative correlation for longer ones ($\ge 100$ words).
This suggests that answer length explains some variance in question salience, but cannot fully account for it.
A hypothesis is that model-generated summaries are abstractive, possibly conveying information more densely than the reference answers.

\section{Pilot Study: Human Salience in News Summaries}
\label{sec:appendix-human-salience-news}
To test the generality of our framework, we use it as a tool to analyze the salience notions encoded in human-written summaries from a standard summarization benchmark. We focus on news articles from the \textsc{cnn/dm} dataset~\cite{Hermann:2015:NIPS}.

\paragraph{Method.} First, we run question generation over a sample of 200 random documents to identify QUDs for this domain (see Steps 1 and 2 in~\cref{sec:method-questions}).
Next, we use the resulting questions to analyze human summaries (see Steps 3 and 4 in~\cref{sec:method-questions}).
Since each article in \textsc{cnn/dm} has only one reference summary, we select \texttt{\small{(document, summary)}} pairs with summaries approximating our target lengths of $L = \{10, 20, 50, 100, 200\}$ words, allowing for a delta of $\pm 10\%$.\footnote{
Of the 287,113 documents in the \textsc{cnn/dm} training set, 0.28/1.60/26.84/3.50/0.04\% fall into the respective buckets.
}
Finally, from each length bucket, we draw a random sample of 200 \texttt{\small{(document, summary)}} pairs for analysis.

\begin{figure*}[t]
\includegraphics[width=\textwidth]{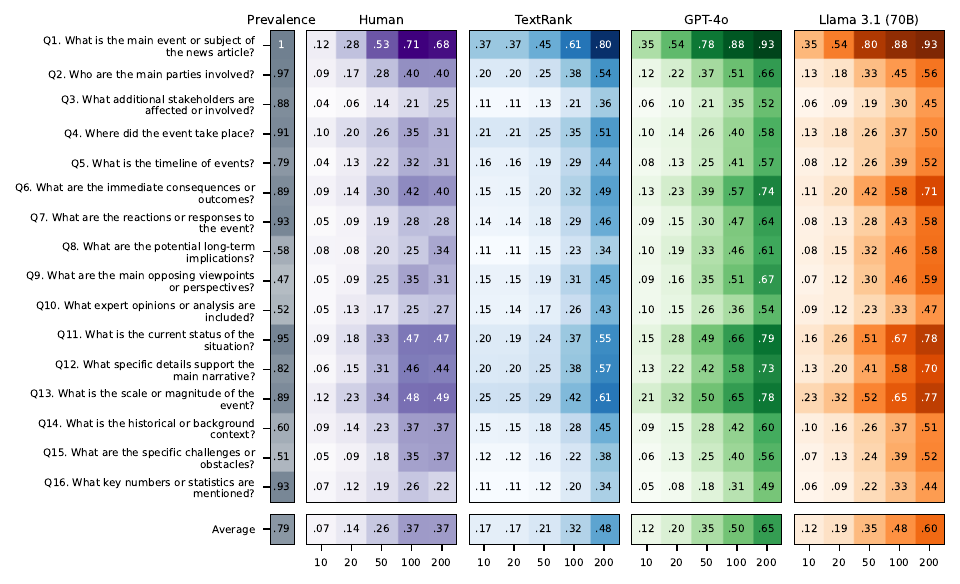}
\caption{Corpus-level content salience map for \emph{CNN/DM}. Salience scores for human summaries and model summaries are derived form different document samples, so cannot be directly compared.}
\label{fig:salience-cnndm-human}
\end{figure*}

\textbf{Results.} We present the content salience map for both human and model summaries in \cref{fig:salience-cnndm-human}. We observe consistent trends in question salience. For example, questions about the main event (\emph{Q1}) or its magnitude (\emph{Q13}) and consequences (\emph{Q6}) consistently achieve higher scores than more detailed questions about reactions (\emph{Q7}), expert opinions (\emph{Q10}) or additional stakeholders (\emph{Q3}). While the salience scores of human and model summaries are not directly comparable due to differing document samples, they exhibit similar trends.

In sum, this pilot study demonstrates the versatility of the framework, and suggests that it could be used in future work to understand human notions of salience on a larger scale.

\section{Responsible NLP Considerations}
\textbf{Compute Requirements.} Experiments were conducted on NVIDIA A100 80GB GPUs, requiring approximately 20 GPU hours per dataset, and an additional 360 GPU hours for the temperature sweep on the RCT dataset, totaling 440 GPU hours.
We ran inference using \textsc{vllm} ({\small\rurl{docs.vllm.ai}}).
GPT-4o models were accessed through the OpenAI API with inference costs $\leq 100\$$.

\textbf{Salience Annotation Study.} Participants joined on a volunteer basis, gave informed consent and agreed that their annotations will be shared in anonymized form in the paper repository. According to our institutional policies, this study did not require institutional review board (IRB) approval.

\textbf{Data Licensing.}
We obtain RCT abstracts in accordance with fair use principles through the PubMed Entrez API.\unfootnote{All URLs accessed 2025-05-15.}\footnote{\scriptsize \rurl{www.ncbi.nlm.nih.gov/home/develop/api/}}
Related work sections of CL and Astro papers were collected via the arXiv API.\footnote{\scriptsize \rurl{info.arxiv.org/help/api/index.html}} While the majority of papers on arXiv is published under the arXiv license\footnote{\scriptsize \rurl{arxiv.org/licenses/nonexclusive-distrib/1.0/license.html}} retaining \emph{copyright} with the original author(s), the \emph{use} of paper contents for research is explicitly granted and encouraged in the arXiv API terms \& conditions.\footnote{\scriptsize \rurl{info.arxiv.org/help/api/tou.html}}
We reused meeting transcripts from QMSum~\cite{Zhong:2021:NAACL}.\footnote{\scriptsize \rurl{github.com/Yale-LILY/QMSum}} All meeting transcripts are under an open use license, such as CC BY 4.0 (academic meetings and product meetings) or Open Government License Version 3 (parliament commitee meetings).\footnote{\scriptsize \rurl{creativecommons.org/licenses/by/4.0/legalcode}}\textsuperscript{,}\footnote{\scriptsize \rurl{groups.inf.ed.ac.uk/ami/icsi/license.shtml}}\textsuperscript{,}\footnote{\scriptsize \rurl{groups.inf.ed.ac.uk/ami/corpus/license.shtml}}\textsuperscript{,}\footnote{\scriptsize \rurl{www.nationalarchives.gov.uk/doc/open-government-licence/}}

\begin{table*}[t]
\sisetup{
detect-weight,
mode=text,
table-format=-1.2,
table-space-text-post={$^{**}$}
}
\small
\centering
\setlength{\tabcolsep}{4pt}
\begin{tabular}{llSSSSSSSS}
\toprule
\B {Measure} & \B {Dataset} & \B {Random} & \B {OLMo} & \B {Mixtral} & \B Llama$^{3.1}_{8b}$ & \B Llama$^{3.1}_{70b}$ & \B {4o-mini} & \B {4o} & \B {Average} \\
\midrule
\multicolumn{10}{c}{\texttt{Consistency of Salience Estimates}}\\\midrule
\multirow[c]{5}{0.14\textwidth}{\emph{LLM-perceived}} & RCT & -0.06 & 0.34$^{*}$ & 0.48$^{**}$ & 0.29$^{*}$ & 0.61$^{**}$ & 0.75$^{**}$ & \B 0.80$^{**}$ & 0.46$^{**}$ \\
 & Astro & 0.02 & 0.07 & 0.41$^{*}$ & 0.56$^{*}$ & 0.64$^{**}$ & 0.76$^{**}$ & \B 0.86$^{**}$ & 0.47$^{**}$ \\
 & CL & -0.10 & -0.05 & 0.65$^{**}$ & 0.29 & \B 0.73$^{**}$ & 0.70$^{**}$ & 0.57$^{**}$ & 0.40$^{**}$ \\
 & QMsum & -0.07 & 0.42$^{**}$ & 0.61$^{*}$ & 0.36 & \B 0.87$^{**}$ & 0.72$^{**}$ & 0.80$^{**}$ & 0.53$^{**}$ \\\cmidrule{2-10}
 & Average & -0.05 & 0.20$^{*}$ & 0.54$^{**}$ & 0.37$^{*}$ & 0.71$^{**}$ & 0.73$^{**}$ & \B 0.76$^{**}$ & 0.47$^{**}$ \\\midrule
\multirow[c]{5}{0.14\textwidth}{\emph{LLM-observed}} & RCT & 0.94$^{**}$ & 0.98$^{**}$ & \B 0.99$^{**}$ & \B 0.99$^{**}$ & \B 0.99$^{**}$ & \B 0.99$^{**}$ & \B 0.99$^{**}$ & \B 0.99$^{**}$ \\
 & Astro & 0.91$^{**}$ & 0.99$^{**}$ & \B 1.00$^{**}$ & 0.97$^{**}$ & 0.99$^{**}$ & 0.97$^{**}$ & 0.97$^{**}$ & 0.97$^{**}$ \\
 & CL & 0.96$^{**}$ & \B 0.99$^{**}$ & \B 0.99$^{**}$ & 0.96$^{**}$ & \B 0.99$^{**}$ & \B 0.99$^{**}$ & 0.97$^{**}$ & 0.98$^{**}$ \\
 & QMsum$^\dagger$ & 0.87$^{**}$ & {---} & 0.99$^{**}$ & 0.99$^{**}$ & \B 1.00$^{**}$ & 0.99$^{**}$ & {---} & 0.97$^{**}$ \\\cmidrule{2-10}
 & Average & 0.92$^{**}$ & \B 0.99$^{**}$ & \B 0.99$^{**}$ & 0.98$^{**}$ & \B 0.99$^{**}$ & 0.98$^{**}$ & 0.98$^{**}$ & 0.97$^{**}$ \\\midrule
\multicolumn{10}{c}{\texttt{Correlation of Salience Estimates}} \\\midrule
 \multirow[c]{5}{0.14\textwidth}{\emph{LLM-perceived vs. LLM-observed}} & RCT & -0.06 & 0.10 & 0.25 & 0.25 & 0.37$^{**}$ & 0.41$^{**}$ & \B 0.51$^{**}$ & 0.28$^{*}$ \\
 & Astro & 0.11 & 0.09 & 0.31 & 0.56$^{**}$ & 0.50$^{*}$ & \B 0.65$^{**}$ & 0.58$^{*}$ & 0.40$^{**}$ \\
 & CL & -0.08 & 0.16 & 0.44 & 0.47$^{*}$ & 0.38 & \B 0.58$^{*}$ & 0.41 & 0.34 \\
 & QMsum$^\dagger$ & 0.11 & {---} & 0.46$^{*}$ & 0.16 & \B 0.63$^{*}$ & 0.60$^{*}$ & {---} & 0.39$^{*}$ \\\cmidrule{2-10}
 & Average & 0.03 & 0.12 & 0.37$^{*}$ & 0.36$^{**}$ & 0.47$^{*}$ & \B 0.56$^{**}$ & 0.50$^{*}$ & 0.35$^{*}$ \\\midrule
\multicolumn{10}{c}{\texttt{Correlation of Model and Human Salience}} \\\midrule
 \multirow[c]{5}{0.14\textwidth}{\emph{LLM-perceived vs. Human}} & RCT & -0.03 & 0.22 & 0.38$^{**}$ & 0.34$^{*}$ & 0.49$^{**}$ & 0.48$^{*}$ & \B 0.56$^{**}$ & 0.35$^{**}$ \\
 & Astro & 0.07 & 0.12 & 0.30$^{**}$ & 0.31$^{*}$ & 0.27 & \B 0.45$^{**}$ & 0.44$^{*}$ & 0.28$^{**}$ \\
 & CL & 0.06 & -0.03 & 0.41$^{*}$ & 0.22 & \B 0.48$^{*}$ & 0.44$^{*}$ & 0.46$^{**}$ & 0.29$^{**}$ \\
 & QMsum & 0.14 & 0.34 & 0.54$^{*}$ & 0.36$^{*}$ & 0.62$^{*}$ & \B 0.67$^{**}$ & \B 0.67$^{**}$ & 0.48$^{*}$ \\\cmidrule{2-10}
 & Average & 0.07 & 0.16 & 0.41$^{*}$ & 0.31$^{*}$ & 0.46$^{**}$ & 0.51$^{**}$ & \B 0.53$^{**}$ & 0.35$^{**}$ \\\midrule
\multirow[c]{5}{0.14\textwidth}{\emph{LLM-observed vs. Human}} & RCT & 0.31 & 0.28 & 0.27 & 0.25 & 0.25 & \B 0.34 & 0.24 & 0.27 \\
 & Astro & 0.11 & 0.25$^{*}$ & 0.27$^{*}$ & 0.29$^{*}$ & \B 0.31 & 0.26 & 0.25$^{*}$ & 0.25$^{*}$ \\
 & CL & \B 0.30 & 0.23 & 0.23 & 0.24 & 0.26 & 0.25 & 0.24 & 0.25 \\
 & QMsum$^\dagger$ & 0.16 & {---} & 0.53$^{*}$ & 0.58$^{**}$ & \B 0.59$^{**}$ & 0.51$^{**}$ & {---} & 0.48$^{*}$ \\\cmidrule{2-10}
 & Average & 0.20 & 0.25 & 0.33$^{*}$ & 0.35$^{*}$ & \B 0.36$^{*}$ & 0.34$^{*}$ & 0.25 & 0.30$^{*}$ \\
\bottomrule
\end{tabular}
\caption{
Spearman rank correlations between salience estimates, split by dataset. Significance: $^{*}$ ($p < 0.05$) and $^{**}$ ($p < 0.01$); row-wise maxima are bolded. $^\dagger$Results for QMSum not available due to limited context window (OLMo) and budget constraints (GPT-4o).}
\label{tab:results-rater-agreement-full}
\end{table*}

\begin{figure*}[t]
\centering

    \begin{subfigure}[b]{\textwidth}
    \includegraphics[width=\textwidth]{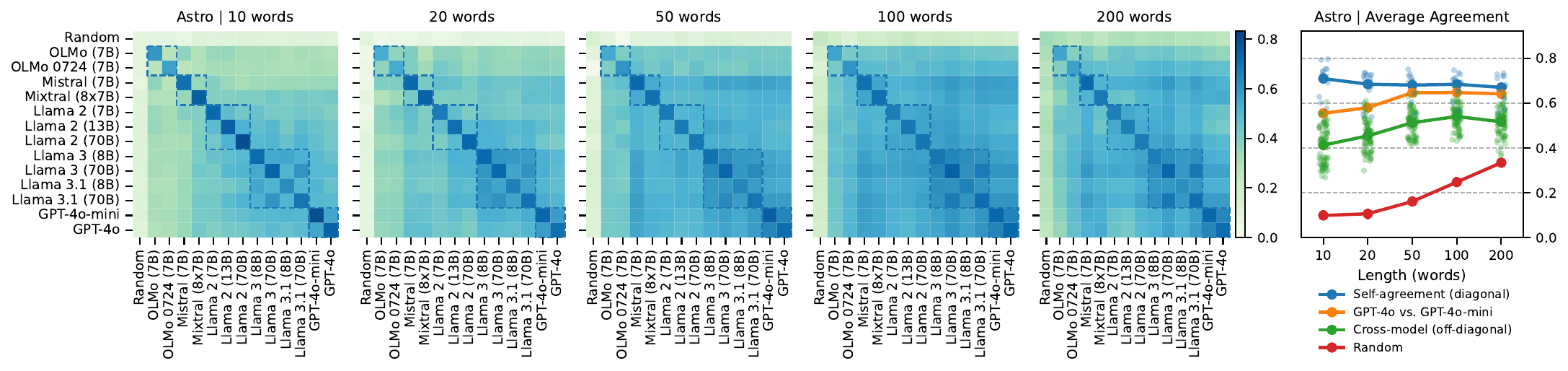}
    \caption{Model similarity for \emph{Astro}.}
    \label{fig:agg-astro}
    \end{subfigure}

    \vspace{0.5cm}

    \begin{subfigure}[b]{\textwidth}
    \includegraphics[width=\textwidth]{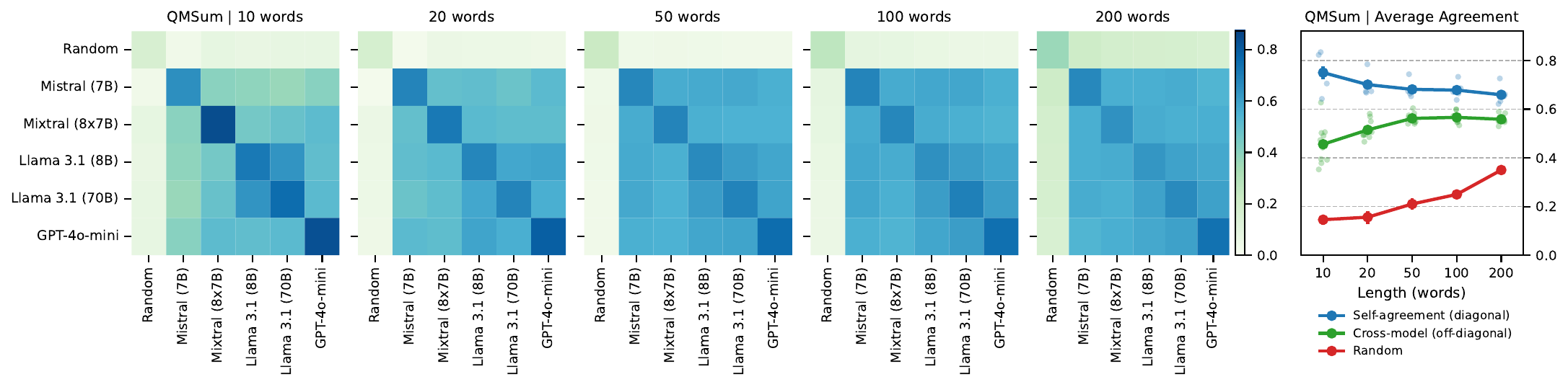}
    \caption{Model similarity for \emph{QMSum}.}
    \label{fig:agg-qmsum}
    \end{subfigure}

\caption{Do models share a similar notion of salience? Heatmaps show agreement on topic inclusion by summary length (Krippendorff's alpha calculated column-wise for the CSM pairs).}
\label{fig:agg-astro-qmsum}
\end{figure*}

\begin{table*}[t]
    \fontsize{8.5pt}{8.5pt}\selectfont
    \begin{tabular}{@{}p{0.17\textwidth}rp{0.23\textwidth}p{0.44\textwidth}@{}}
    \toprule
        \textbf{Question} &
        $L_o$/$L_p$/$H$ &
        \faRobot\ \textbf{Rationale LLM-perceived} &
        \faUsers\ \textbf{Rationales Human}\\
    \midrule
    \multicolumn{4}{c}{\emph{continued from \cref{tab:results-examples}}} \\
    \textbf{(\#4, RCT \faUserMd)}
        What were the comparative results between intervention and control groups? & 3.2/4.6/5.0 &
        Comparative results are key to understanding the effectiveness of the intervention. &
        (H\textsubscript{\{1,2,3\}} = 5) Brief comparison between groups allows for readers to quickly assess statistical significance; Comparative results are one of the primary outputs reflecting effectiveness; essential for understanding the findings. \\

    \textbf{(\#5, RCT \faUserMd)}
        What implications or future recommendations did the study suggest based on its findings? & 4.6/3.6/2.0 &
        Implications and recommendations are important for understanding the study's impact and future research directions. &
        (H\textsubscript{1} = 1) This is a job for the reader to decide for himself or herself; (H\textsubscript{2} = 3) helps the reader understand the paper. Useful for at the end of a summary. (H\textsubscript{3} = 2) While helpful, they are not necessarily always based on the opinion of the authors; would have the reader get to that conclusion themselves.\\

    \textbf{(\#6, RCT \faUserMd)}
        What is the main outcome or effect observed? & 4.6/5.0/5.0 &
        The main outcome is critical to understanding the study's primary conclusion. &
        (H\textsubscript{\{1,2,3\}} = 5) determines whether a paper can be used to influence decision making; primary information when reading summary \\

    \textbf{(\#7, RCT \faUserMd)}
        How were the participants or subjects of the study selected and divided? & 3.7/3.2/2.3 &
        Participant selection and division are important for assessing the study's validity. &
        (H\textsubscript{1} = 2) This helps a reader determine external validity of a study;
        (H\textsubscript{2} = 1) Not necessary as fairly standardized;
        (H\textsubscript{3} = 4) Crucial to evaluate the study's validity and relevance \\

    \textbf{(\#8, Astro \faStar)}
        What limitations or considerations are noted by the study? & 1.1/3.6/2.3 &
        Limitations are crucial for interpreting the study's reliability and potential biases. &
        (H\textsubscript{1} = 3) important but best reserved for readers who choose to critically examine the study in its totality; (H\textsubscript{2} = 1) Typically addressed in the discussion section of a paper, but is not necessary for a summary; (H\textsubscript{3} = 3) would note if I had the space, however, I would also implore the readers to think about limitations themselves. \\

    \textbf{(\#9, Astro \faStar)}
        How do the findings relate to existing models or theories? & 1.6/4.0/3.8 &
        This is important for understanding the study's impact on current scientific understanding. &
        (H\textsubscript{1} = 4) I think this is important to describe the finding in context of the research question, initial hypothesis, or a theory.  (H\textsubscript{2} = 3) This is important for the broader context of the study, however it is more important to present the study results in a clear way. Only after that is clear I'd compare to other studies.(H\textsubscript{3} = 5) This forms the main body of a discussion section. (H\textsubscript{4} = 3) Situating the findings within the literature by relating to existing models or theories is important.(H\textsubscript{3} = 4) Comparing with existing literature is very important for the discussion section.\\

    \textbf{(\#10, Astro \faStar)}
        What are the main findings of the study? & 3.5/5.0/4.8 &
        The main findings are the core of the study and must be included in any summary. &
        (H\textsubscript{1} = 5) key takeaway from the paper and should be included regardless of what the paper is about (H\textsubscript{2} = 5) most important information of the summary (H\textsubscript{3} = 4) The main findings should be briefly addressed in the summary of a discussion for the reader's quick follow-up (H\textsubscript{4} = 5) The main findings of the study, along with the main focus, form the two most important elements of an article summary. (H\textsubscript{5} = 5) Important to state the main findings and then discuss them in details. \\

    \textbf{(\#11, Astro \faStar)}
        What specific challenges or limitations does the study address or identify? & 1.6/3.2/2.6 &
        Understanding the challenges or limitations provides context for the study's reliability and areas for improvement. &
        (H\textsubscript{1} = 1) I most likely do not include challenges and limitations. These examples focused on the future needs not an existing open question. The focus will be on the findings in the context of a hypothesis, conjecture, or a theory. (H\textsubscript{2} = 1) Level of detail that a reader would need only if interested in full paper. Some challenges can be identified if the methods and scope of the paper are summarized clearly. (H\textsubscript{3} = 5) This forms the main body of a discussion section. (H\textsubscript{4} = 2) depends upon the significance of those challenges or limitations (H\textsubscript{5} = 4) Identify the limitations and challenges of the study is very important \\
    \bottomrule
    \end{tabular}
    \caption{Example questions, salience scores by LLM-observed ($L_o$, rescaled to 1-5), LLM-perceived ($L_p$), humans ($H$) and summarized rationales.}
    \label{tab:results-examples-part2}
\end{table*}

\clearpage
\onecolumn
\section{LLM Prompts}
\label{sec:appendix-prompts}
This section provides all prompts used throughout the experiments. Summarization (\cref{lst:summarization,lst:summarization-meetings}), question generation (\cref{lst:qg}), question answering (\cref{lst:qa}), answer claim  splitting (\cref{lst:claim-split}), and introspection (\cref{lst:introspection}).

\codeboxinput[label=lst:summarization]{Summarization prompt}{prompts/summarization.txt}
\codeboxinput[label=lst:summarization-meetings]{Summarization prompt for meeting transcripts}{prompts/summarization-meetings.txt}
\codeboxinput[label=lst:qg]{Question generation prompt}{prompts/qg.txt}
\codeboxinput[label=lst:qa]{Question answering prompt}{prompts/qa.txt}
\codeboxinput[label=lst:claim-split]{Claim splitting prompt}{prompts/claim-splitting.txt}
\codeboxinput[label=lst:introspection]{Introspection prompt}{prompts/introspection.txt}

\twocolumn
\section{Question Salience Annotation Guidelines}
\label{sec:appendix-annotation-guidelines}

\paragraph{Motivation.}
When summarizing long texts, we must consciously decide what information to include or exclude from a summary. These decisions are grounded in a notion of information salience, or how important we consider the information for our intended audience. We study this phenomenon in the context of automatic text summarization systems. Specifically, we aim to understand how well these systems replicate the judgments of domain experts regarding what information is most relevant.\looseness=-1

\paragraph{Task.}
Imagine you are asked to \textbf{summarize a paper describing the results of a randomized controlled trial (RCT)} for a typical reader in this field. The summary should provide enough context to stand alone, since the reader will only see your summary and no other parts of the paper. Furthermore, the summary length is constrained, requiring you to think about what content to prioritize. In this study, we frame content as questions that a summary could answer.

Ask yourself: \textbf{What are some key questions you want the summary to answer?} Your task is to rate the relative importance of a list of questions on the following scale.
\begin{todolist}[noitemsep]
\item (1) Least important; I would exclude this information from a summary.
\item (2) Low importance; I would include this information if there is room.
\item (3) Medium importance; I would probably include this information.
\item (4) High importance; I would definitely include this information.
\item (5) Most important; One of the first questions to be answered in the summary.
\end{todolist}
\paragraph{Rationale.}
For each rating, please provide a brief (1-sentence) rationale explaining your decision or highlighting any considerations or uncertainties.

\paragraph{Example answers.}
To give you a feeling for the kind of content a question might elicit, all questions have an illustrative answer sourced from a randomly chosen document (= RCT paper). Please keep the following in mind:
\begin{itemize}[noitemsep]
    \item \emph{Answer length} does not determine the question's importance.
    \item \emph{Phrasing and selection.} The precise answer phrasing can be different in the summary, and not all answer content must appear in the summary.
    \item \emph{Overlap.} Some questions may elicit overlapping answers. Therefore, focus on the essence of each question. Remember that in an actual summary, overlapping answer information would only be stated once, so don't worry about it (see below).
    \item \emph{Relevance.} The questions are answerable with most documents in this genre. Do your rating on the assumption that the document talks about this information.
\end{itemize}
\paragraph{Suggested process.}
\begin{enumerate}[noitemsep]
    \item Read all questions first.
    \item Identify questions that seem most/least important, and rate these as “anchor points.”
    \item Then, rate the remaining questions.
\end{enumerate}
Finally, there are no right or wrong ratings. Use your best judgment and intuition. Thank you for participating!

\subsection*{Appendix: Example of overlapping answers.}
Consider questions \texttt{Q1-Q3} below. Each question asks for a distinct unit of information, but the answer of \texttt{Q3} overlaps with the answer of \texttt{Q1} and \texttt{Q2}. The overlapping information is highlighted in \hlorange{orange} while the \emph{essence of the question} is highlighted in \hlgreen{green}. Base your rating on the essence of the question.

\begin{subbox}[width=1\linewidth, center]{Example of overlapping answers}
\small
\textbf{Q1. What was the study design or setting of the trial?}
This trial is a multicentre, randomized, double-blind, phase 3 study.\\

\textbf{Q2. What specific treatments were compared in the study?}
DBPR108 100 mg, sitagliptin 100 mg, and placebo.\\

\textbf{Q3. How were the participants or subjects of the study selected and divided?}
In this \hlorange{multicentre, randomized, double-blind, phase 3 study}, adult patients with type 2 diabetes were \hlgreen{randomly assigned} to \hlorange{receive either DBPR108 100mg, sitagliptin 100mg, or placebo} once daily. \hlgreen{A total of 766 patients were enrolled and divided into three groups: DBPR108 100mg (n=462), sitagliptin 100mg (n=152), or placebo (n=152).}
\end{subbox}

\begin{figure*}
\includegraphics[width=\textwidth]{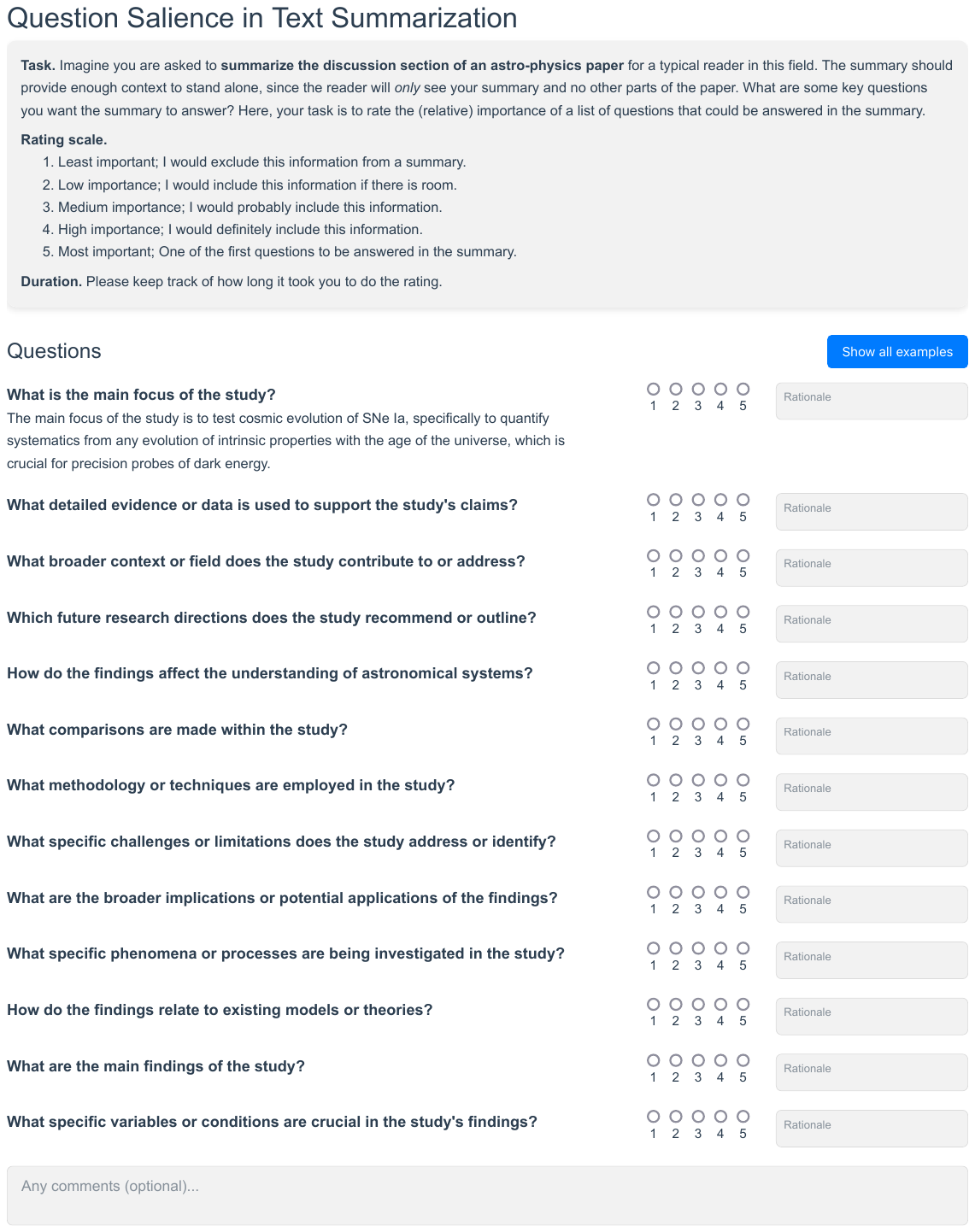}
\caption{Interface for question salience annotation. Each question can be expanded to show an illustrative answer sourced from a randomly chosen document. The questions shown here are for the \emph{Astro} dataset.}
\end{figure*}

\end{document}